\documentclass[12pt]{article}
\usepackage{enumerate}
\usepackage{graphicx}
\RequirePackage{natbib}
\usepackage{smile}
\usepackage{amsmath}
\usepackage{amsthm}
\usepackage{amssymb}
\usepackage{algorithm}
\usepackage{setspace}
\usepackage{soul}
\usepackage{color}
\usepackage[left=2.54cm,top=2.54cm,right=2.54cm,bottom=2.54cm]{geometry}
\newcommand\independent{\protect\mathpalette{\protect\independenT}{\perp}}
\def\independenT#1#2{\mathrel{\rlap{$#1#2$}\mkern2mu{#1#2}}}
\usepackage{fullpage}

\def\T{\mathrm{\scriptstyle T}} %%%transpose operator
\usepackage[colorlinks,
linkcolor=red,
anchorcolor=blue,
citecolor=blue
]{hyperref}

\begin{document}

\title{A convex formulation for high-dimensional sparse sliced inverse regression}
\author{Kean Ming Tan, Zhaoran Wang, Tong Zhang, Han Liu, and R. Dennis Cook}
\maketitle

\begin{abstract}
Sliced inverse regression is a popular tool for sufficient dimension reduction, which replaces covariates with a minimal set of their linear combinations without loss of information on the conditional distribution of the response given the covariates. The estimated linear combinations include all covariates, making results difficult to interpret and perhaps unnecessarily variable, particularly when the number of covariates is large.
In this paper, we propose a convex formulation for fitting sparse sliced inverse regression in high dimensions.  Our proposal estimates the subspace of the linear combinations of the covariates directly and performs variable selection simultaneously.  We solve the resulting convex optimization problem via the linearized alternating direction methods of multiplier algorithm, and establish an upper bound on the subspace distance between the estimated and the true subspaces.  
Through numerical studies, we show that our proposal is able to identify the correct covariates in the high-dimensional setting. 
\end{abstract}

\section{Introduction}

\label{sec:intro}
We consider regression of a univariate response $y\in \RR$ on a  stochastic covariate vector $x=(x_1,\ldots,x_d)^\T\in \RR^{d}$ in which the number of covariates $d$ exceeds the sample size $n$.  The goal is to infer the conditional distribution of $y$ given $x$.
When $d$ is large, it is often desirable to  perform dimension reduction on the covariates with the aim of minimizing information loss.
Sufficient dimension reduction is popular for this purpose \citep{li1991sliced,cook1994interpretation,cook2009regression}. 

Let $K < \min (n,d)$ and let $\beta_1,\ldots,\beta_K\in \RR^{d}$ be $d$-dimensional vectors.  We assume that 
 \begin{equation}
 \label{Eq:conditionally independent}
y \independent x \mid \left( \beta_1^\T x,\ldots,\beta_K^\T x  \right),
 \end{equation} 
where $\independent$ signifies independence.
Equation~\eqref{Eq:conditionally independent} implies that $y$ can be explained by a set of $K$ linear combinations of $x$. 
A dimension reduction subspace $\cV$ is defined as the subspace spanned by $\beta_1,\ldots,\beta_K$ such that (\ref{Eq:conditionally independent}) holds.  
We henceforth refer to $\beta_1,\ldots,\beta_K$ as the sufficient dimension reduction directions.  
Dimension reduction subspaces are not unique in general, and
\citet{cook1994interpretation} defined the  central subspace, $\cV_{y\mid x}$, as the intersection of all dimension reduction subspaces. 
Under regularity conditions, the central subspace exists and is also the unique minimum dimension reduction subspace that satisfies (\ref{Eq:conditionally independent}).
Many authors have proposed methods to estimate the central subspace \citep{li1991sliced,cook1991comment,cook1999dimension,bura2001extending,bura2001estimating,dennis2000save,cook2007fisher,cook2008principal,li2007directional,cook2009likelihood,ma2012semiparametric,ma2013efficient}.
The sufficient dimension reduction literature is vast: see \citet{ma2013review} for a comprehensive list of references.

We focus on sliced inverse regression for estimating the central subspace $\cV_{y \mid x}$ \citep{li1991sliced}. 
In the low-dimensional setting in which $d< n$,
 the central subspace $\cV_{y \mid x}$ can be estimated consistently \citep{li1991sliced,hsing1992asymptotic,zhu1995asymptotics,zhu1996asymptotics,zhu2006sliced}.   
%\cred -- deleted sentence since it is stated at the end of the first paragraph of section 2 -- If we further assume that the conditional distribution of $x$ given $y$ is normally distributed and $y$ is categorical, then sliced inverse regression  gives the maximum likelihood estimator of the central subspace (\citealp{cook2008principal}, Section 4.1).  
One drawback of sliced inverse regression  is that the estimated sufficient dimension  reduction directions involve all $d$ covariates, so these directions are hard to interpret, and important covariates may be difficult to identify.  

Numerous attempts have been made to perform variable selection for sliced inverse regression in the low-dimensional setting  \citep{cook2004testing,li2005model,ni2005note,li2008sliced,li2007sparse}.
Most are conducted stepwise, estimating a sparse solution for each direction.  
However, sparsity in each sufficient dimension reduction direction does not correspond to variable selection unless an entire row of the basis matrix $(\beta_1,\ldots,\beta_K)$
%deleted $\beta$ since it is undefined 
is set to zero, and \citet{chen2010coordinate} proposed a novel penalty to encourage this.  Their proposal involves solving a non-convex problem and a global optimum solution is often not guaranteed.

In the high-dimensional setting, \citet{linetal2017} proposed a screening approach to perform variable selection. The selected variables are then used to fit classical sliced inverse regression.  
\citet{hilafu} proposed a sequential approach for estimating high-dimensional sliced inverse regression.  
Both proposals are step-wise procedures that do not correspond to solving a convex optimization problem.  
Moreover, as discussed in \citet{hilafu}, theoretical properties for their proposed estimators are hard to establish due to the sequential procedure used to obtain the estimators.    

\citet{yu2013dimension} proposed using $\ell_1$-minimization with an adaptive Dantzig selector, and established a non-asymptotic error bound for the resulting estimator.  \citet{wang2015estimating} recast sliced inverse regression as a reduced-rank regression problem, proposed solving a  non-convex optimization problem for simultaneous variable selection and dimension reduction, and showed that their proposed method is prediction consistent.  However, there is a gap between the optimization problem and the theoretical results: there is no guarantee that the estimator obtained from solving the proposed biconvex optimization problem is the global minimum.

Most existing work in the high-dimensional sufficient dimension reduction literature involves non-convex optimization problems. Moreover, they seek to estimate a set of reduced predictors that are not identifiable by definition, rather than the central subspace.  
In this paper, we propose a convex formulation for sparse sliced inverse regression in the high-dimensional setting by adapting techniques from sparse canonical correlation analysis  \citep{vu2013fantope,gao2014sparse}.  
Our proposal estimates the central subspace directly and performs variable selection simultaneously.  
Moreover, the proposed method can be adapted for sufficient dimension reduction methods that can be formulated as generalized eigenvalue problems.  These include sliced average variance estimation, directional regression, principal fitted components, principal hessian direction, and iterative hessian transformation.

%%%%%%%%%%%%%%%%%%%%
% Setup and Formulation
%%%%%%%%%%%%%%%%%%%%
\section{A Review on Sliced Inverse Regression}
\subsection{Sliced inverse regression}
\label{sec:gep}
\citet{li1991sliced} considered the general regression model
\begin{equation}
\label{Eq:regression}
y = f(\beta_1^\T x,\ldots,\beta_K^\T x,\epsilon),
\end{equation}
where $\epsilon$ is a stochastic error independent of $x$ and $f(\cdot)$ is an unknown link function.  
Model (\ref{Eq:regression}) is equivalent to (\ref{Eq:conditionally independent}) in the sense that 
the conditional distribution of $y$ given $x$ is captured by a set of $K$ linear combinations of $x$ \citep[][Lemma 1]{zeng2010integral}.
It has been shown that the central subspace 
$\cV_{y\mid x}$ spanned by  $\beta_1,\ldots,\beta_K$ can be identified. 
In fact, sliced inverse regression gives the maximum likelihood estimator of the central subspace if $x$ given $y$ is normally distributed and $y$ is categorical \citep[][\S 4.1]{cook2008principal}.

Sliced inverse regression requires the linearity condition on the covariates $x$: for any $a \in \RR^d$,
%\cred -- deleted phrase since it's redundant -- the conditional expectation is linear in $\beta_1^\T x,\ldots,\beta^\T_K x$,
\begin{equation}
\label{Eq:linearity}
E(a^\T x \mid \beta_1^\T x,\ldots,\beta_K^\T x) =  b_0 + b_1 \beta_1^\T x + \cdots+b_K \beta_K ^\T x
\end{equation}
for some constants $b_0,\ldots,b_K$. The linearity condition (\ref{Eq:linearity}) is satisfied when the distribution of $x$ is elliptically symmetric \citep{li1991sliced}.  For instance, (\ref{Eq:linearity}) holds when $x$ is normally distributed with covariance matrix $\Sigma_x$.
The linearity condition involves only the marginal distribution of $x$ and is regarded as mild in the sufficient dimension reduction literature.

Under the linearity condition (\ref{Eq:linearity}), the inverse regression curve 
 $E(x \mid y)$ resides in the linear subspace spanned by $\Sigma_x \beta_1,\ldots,\Sigma_x \beta_K$ \citep[][Theorem 3.1]{li1991sliced}. In other words, $\Sigma_{E(x\mid y)} \beta_k = \lambda_k \Sigma_x\beta_k$ for $k=1,\ldots,K$, where  $\Sigma_{E(x\mid y)}$ is the covariance matrix of the conditional expectation $E(x\mid y)$, $\lambda_k$ is the $k$th largest generalized eigenvalue, $\beta_k^\T \Sigma_x \beta_k=1$ and $\beta_j^\T \Sigma_x \beta_k = 0$ for $j\ne k$.  Let the columns of $V \in \RR^{d\times K}$ represent a basis for $\cV_{y\mid x}$.  Then  a basis can be estimated by solving the generalized eigenvalue problem
% Thus, using a matrix representation of $\beta_1,\ldots,\beta_K$, a basis of the central subspace can be estimated by solving the generalized eigenvalue problem 
\begin{equation}
\label{eq:1}
\hat{\Sigma}_{E(x\mid y)}  V =  \hat{\Sigma}_x V \Lambda,
\end{equation}
where $\hat{\Sigma}_{E(x\mid y)}$ is an estimator of  $\Sigma_{E(x\mid y)}$, $V \in \RR^{d\times K}$ consists of $K$ eigenvectors  such that $V^\T \hat{\Sigma}_x V = I_K$, and $\Lambda = \mathrm{diag}(\lambda_1,\ldots,\lambda_K) \in \RR^{K\times K}$.  
By definition, $\Sigma_{E(x \mid y)}$ is of rank $K$.
  An estimator of $V$ can be obtained equivalently by solving the non-convex optimization problem
\begin{equation}
\label{Eq:variational}
\underset{V \in \RR^{d\times K}}{\mathrm{minimize}} \; -\mathrm{tr}\left\{V^\T \hat{\Sigma}_{E(x\mid  y)} V\right\} \; \text{subject~to~} V^\T \hat{\Sigma}_x V = I_K.
\end{equation}
Let $\hat{V}$ be a solution of (\ref{Eq:variational}).  Then, the central subspace is estimated as $\mathrm{span}(\hat{V})$ and  the sufficient dimension reduced variables are $\hat{V}^\T x$.

%%%%%%%%%%%%%%%%%%%%%%%%%%%%%%%%%%
% Estimators for the conditional covariance
%%%%%%%%%%%%%%%%%%%%%%%%%%%%%%%%%%
\subsection{Estimators for the conditional covariance}
\label{subsec:problem set up}
Let $(y_1,x_1),\ldots,(y_n,x_n)$ be $n$ independent and identically distributed observations.  
We denote the order statistics of the response by $y_{(1)}\le \cdots \le y_{(n)}$. 
In addition, define $x_{(i)^*}$ as the value of $x$  associated with the $i$th order statistic of $y$. 
For instance, if the fifth observation $y_5$ is the largest then $y_{(n)}=y_5$ and $x_{(n)^*} = x_5$.

To estimate $\Sigma_{E(x\mid y)}$ we use the identity $\mathrm{cov}\{E(x \mid y)\}  = \mathrm{cov}(x) -  E \{\mathrm{cov}(x \mid y)\}$.
Let $T =   E \{\mathrm{cov}(x \mid y)\}$. Then,  $\hat{\Sigma}_{E(x \mid y)} = \hat{\Sigma}_x-\hat{T}$, 
where $\hat{\Sigma}_x$ is the sample covariance matrix of $x$ and $\hat{T}$ is an estimator of $T$.
There are two widely used estimators for $T$.  
The first is
\begin{equation}
\label{Eq:estimator conditional expectation}
\hat{T} = \frac{1}{n} \sum_{i=1}^{\lfloor n/2\rfloor} \{x_{(2i)^*}-x_{(2i-1)^*}\}\{x_{(2i)^*}-x_{(2i-1)^*}\}^\T,
\end{equation}
where $\lfloor n/2 \rfloor$ denotes the largest integer less than or equal to $n/2$. 
%\citet{hsing1992asymptotic} showed that the estimator (\ref{Eq:estimator conditional expectation}) is a consistent estimator of $\Tb$ in the low-dimensional setting.
%In Section~\ref{sec:theory}, we will show that $\hat{\Tb}$ is consistent in the high-dimensional setting, under the $\ell_{\infty}$ norm.

The second estimator of $T$ can be obtained by partitioning the $n$ observations into $H$ slices according to the order statistics of $y$ and then computing the weighted average of the sample covariance matrices within each slice.  Let $S_1,\ldots,S_H$ be $H$ sets containing the indices of $y$ partitioned according to their order statistics.   Then,
\begin{equation}
\label{Eq:estimator conditional expectation2}
\tilde{T} = \frac{1}{H} \sum_{h=1}^H \left\{ \frac{1}{n_h} \sum_{i \in S_h} \left( x_i - \bar{x}_{S_h}  \right)\left( x_i - \bar{x}_{S_h}  \right)^\T\right\}.
\end{equation}

Several authors have shown that $\hat{T}$ and $\tilde{T}$ are consistent estimators of $T$ in the low-dimensional setting \citep{hsing1992asymptotic,zhu1995asymptotics,zhu1996asymptotics}. \citet{zhu2006sliced} established consistency for $\tilde{T}$ when $d$ increases as a function of $n$, but at a slower rate than $n$.  \citet{dai2015difference} studied an estimator of the form in \eqref{Eq:estimator conditional expectation} in the context of nonparametric regression.  %See \citet{zhu2006sliced} and \citet{dai2015difference} for a comprehensive list of references.
In \S~\ref{sec:theory}, we will show that $\hat{T}$ converges to $T$ in the high-dimensional setting under the max norm.  Similar results can be shown for $\tilde{T}$.

% ---------------------------------------------------------------------------%
% ---------------------------------------------------------------------------%
%%%%%%%%%%%%%%%%%%%%%%%%%%%%%%
%%%%%%%%%%%%%%%%%%%%%%%%%%%%%%
% Our proposal
%%%%%%%%%%%%%%%%%%%%%%%%%%%%%%
%%%%%%%%%%%%%%%%%%%%%%%%%%%%%%
% ---------------------------------------------------------------------------%
% ---------------------------------------------------------------------------%

\section{Convex Sparse Sliced Inverse Regression}
\label{sec:proposal}
\subsection{Problem formulation}
\label{subsec:proposal}
Recall from \S~\ref{sec:gep} that the goal of sliced inverse regression  is to estimate the central subspace spanned by $\beta_1,\ldots,\beta_K$. 
Thus, instead of estimating each column of $V$ as in (\ref{Eq:variational}), we propose to directly  estimate the orthogonal projection  $\Pi = VV^\T$  onto the subspace spanned by $V$.
By a change of variable, (\ref{Eq:variational}) can be rewritten as
\begin{equation}
\label{Eq:variational1.5}
\underset{\Pi \in\mathcal{M} }{\text{minimize}} \; -\mathrm{tr}\left\{ \hat{{\Sigma}}_{E(x \mid y)} \Pi \right\} \; \text{subject~to~}       \hat{\Sigma}_x^{1/2} \Pi \hat{\Sigma}_x^{1/2} \in \mathcal{B},
\end{equation}
where $\mathcal{B} = \{  \hat{\Sigma}_x^{1/2} \Pi \hat\Sigma_x^{1/2} : V^\T \hat{\Sigma}_x V = I_K    \}$ and $\mathcal{M}$ is the set of $d\times d$ symmetric positive semi-definite matrices.

Instead of solving the non-convex optimization problem in (\ref{Eq:variational1.5}), 
we propose the convex relaxation 
\begin{equation}
\label{Eq:variational2}
\underset{\Pi \in \mathcal{M} }{\text{minimize}} \; -\mathrm{tr}\left\{\hat{\Sigma}_{E(x\mid y)} \Pi\right\} \; \text{subject~to~} \|\hat{\Sigma}_{x}^{1/2} \Pi {\Sigma}_{x}^{1/2}\|_{*}\le K, \; \|\hat{\Sigma}_{x}^{1/2} \Pi \hat{\Sigma}_{x}^{1/2}\|_{\mathrm{sp}}\le 1,
\end{equation}
where 
\begin{eqnarray*}
\|\hat{\Sigma}_{x}^{1/2} \Pi {\Sigma}_{x}^{1/2} \|_* & = & \mathrm{trace}(\hat{\Sigma}_{x}^{1/2} \Pi {\Sigma}_{x}^{1/2}), \\
 \|\hat{\Sigma}_{x}^{1/2} \Pi {\Sigma}_{x}^{1/2}\|_\mathrm{sp} & = & \sup_{v:v^\T v=1}  \left\{\sum_{j=1}^d (\hat{\Sigma}_{x}^{1/2} \Pi {\Sigma}_{x}^{1/2} {v})_j^2 \right\}^{1/2},
\end{eqnarray*}
are the nuclear norm and the spectral norm, respectively.
The nuclear norm constrains the solution to be of low rank
and the spectral norm constrains the maximum eigenvalue of the solution. A similar convex relaxation has been used in sparse principal component analysis and canonical correlation analysis \citep{vu2013fantope,gao2014sparse}. 

To achieve variable selection, we impose a lasso penalty on $\Pi$ to encourage the estimated subspace to be sparse. To this end, we introduce the notion of subspace sparsity. 
%%%%%%%%%%%%%%%%%%%%%%%%%%%%%
% Definition
%%%%%%%%%%%%%%%%%%%%%%%%%%%%%
\begin{definition}
\label{def:subspace sparsity}
Let $\Pi= VV^\T$ be the orthogonal projection matrix onto the subspace $\cV$.  The sparsity level of $\cV$ is the total number of non-zero diagonal elements in $\Pi$, $s = |\mathrm{supp} \{\mathrm{diag} 	(\Pi)\}|$.
\end{definition}
 %We provide an example to clarify Definition~\ref{def:subspace sparsity}.  
Suppose, for example,  that $\Pi_{jj}=0$.  Since $\Pi_{jj} = \sum_{k=1}^K  V_{jk}^2$, this implies that  $V_{jk}= 0$ for all $k\in (1,\ldots,K)$. 
That is, the entire $j$th row of $V$ is zero when $\Pi_{jj} = 0$, which  corresponds to not selecting the $j$th variable.  
It seems intuitive to use the trace penalty to penalize only the diagonal elements of $\Pi$ for variable selection.  However, if a diagonal element of $\Pi$ is zero, the elements in the corresponding row and column of $\Pi$ are zero.  This motivates us to impose an $\ell_1$ penalty on all elements of $\Pi$.

To encourage sparsity, we propose solving the optimization problem
%%replaced \quad with \; three times
 \begin{equation}
 \label{Eq:variational3}
\underset{\Pi \in \mathcal{M}}{\text{minimize}}\; -\mathrm{tr}\left\{\hat{\Sigma}_{E(x \mid y)} \Pi\right\} + \rho  \|\Pi\|_{1} \; \text{subject~to~} \|\hat{\Sigma}_{x}^{1/2} \Pi \hat{\Sigma}_{x}^{1/2}\|_{*}\le K,\; \|\hat{\Sigma}_{x}^{1/2} \Pi \hat{\Sigma}_{x}^{1/2}\|_{\mathrm{sp}}\le 1,
 \end{equation}
where $\|\Pi\|_1 = \sum_{i,j} |\Pi_{ij}|$, and $\rho$ is a positive tuning parameter that controls the sparsity of the solution $\hat{\Pi}$.  Unlike most existing work, our proposal does not require the inversion of 
the empirical covariance matrix $\hat{\Sigma}_{x}$. By Definition~\ref{def:subspace sparsity}, the estimated sparse solution $\hat{\Pi}$ from solving \eqref{Eq:variational3} will yield sparse basis vectors.

%%%%%%%%%%%%%%%%%%%%%%%%%%%%%%
% Algorithm
%%%%%%%%%%%%%%%%%%%%%%%%%%%%%%
\subsection{Linearized alternating direction of method of multipliers algorithm}
\label{subsec:algorithm}
The main difficulty in solving~(\ref{Eq:variational3}) is the interaction between
the penalty term and the constraints.  
To solve (\ref{Eq:variational3}), we use the linearized alternating direction method of multipliers algorithm that allows us to decouple terms that are difficult to optimize jointly \citep{zhang2011unified,wang2012linearized,yang2013linearized}. 
Convergence of the algorithm has been studied in \citet{fang2015generalized}.
The details are presented in Algorithm~\ref{Alg:ssliced inverse regression} and its derivation is deferred to the Appendix. 
Algorithm~\ref{Alg:ssliced inverse regression} amounts to  performing soft-thresholding, computing a singular value decomposition, and modifying the obtained singular values with a monotone piecewise linear function.

Optimization problem (\ref{Eq:variational3}) can also be solved via the standard alternating direction method of multipliers algorithm
 \citep{BoydADMM}. 
In this case, however, there is no closed-form solution for updating the primal variable $\Pi$  as in Step 3(a) of Algorithm~\ref{Alg:ssliced inverse regression}.  Instead of soft-thresholding, it
 involves solving a $d^2$-dimensional lasso regression problem 
in each iteration, which may be computationally prohibitive when the number of covariates $d$ is large.

%%%%%%%%%%%%%%%%%%%%
% Algorithm
%%%%%%%%%%%%%%%%%%%%
\begin{algorithm}[!htp]
\small
\caption{Linearized Alternating Direction of Method of Multipliers Algorithm.}
\label{Alg:ssliced inverse regression}
\begin{enumerate}
\item Input the variables: $\hat{\Sigma}_{x}$, $\hat{\Sigma}_{E(x \mid y)}$, the tuning parameter $\rho$, rank constraint $K$, the L-ADMM parameters $\nu>0$, tolerance level $\epsilon>0$, and $\tau = 4 \nu  \lambda_{\max}^2 (\hat{\Sigma}_{x})$, where $\lambda_{\max}  (\hat{\Sigma}_{x})$ is the largest eigenvalue of $\hat{\Sigma}_x$.
\item  Initialize the parameters: primal variables $\Pi^{(0)} = I_d$, $H^{(0)} = I_d$, and dual variable $\Gamma^{(0)}= {0}$.
 
\item  Iterate until the stopping criterion $\| \Pi^{(t)}-  \Pi^{(t-1)} \|_{\mathrm{F}} \le \epsilon$ is met, where ${\Pi}^{(t)}$ is  $\Pi$ obtained at the $t$th iteration:

\begin{enumerate}
\item $\Pi^{(t+1)}=  \mathrm{Soft}[ \Pi^{(t)}+  \hat{\Sigma}_{E(x \mid y)}/\tau -
\nu \{\hat{\Sigma}_{x} \Pi^{(t)} \hat{\Sigma}_{x} -\hat{\Sigma}_{x}^{1/2}(  H^{(t)
}- \Gamma^{(t)}  ) \hat{\Sigma}_{x}^{1/2}\}/\tau,  \rho/\tau      ]$, where $\mathrm{Soft}$ denotes the soft-thresholding operator, applied element-wise to a matrix,  $\mathrm{Soft}(A_{ij},b) = \text{sign}(A_{ij})  \max\left( |A_{ij}|-b, 0\right)$.

\item $H^{(t+1)} = \sum_{j=1}^d \min \{ 1, \max \left(\omega_j-\gamma^*,0\right)\} u_j u_j^\T$, where $\sum_{j=1}^d \omega_j {u}_j{u}_j^\T$ is the singular value decomposition of  $ \Gamma^{(t)} + \hat{\Sigma}_{x}^{1/2} \Pi^{(t+1)} \hat{\Sigma}_{x}^{1/2}$, and 
\[
\gamma^* = \underset{\gamma>0 }{\mathrm{argmin}}\; \gamma, \qquad \mathrm{subject \; to\;} \sum_{j=1}^d \min \{ 1, \max \left(\omega_j-\gamma,0\right)\}\le K.
\]

\item $\Gamma^{(t+1)} = \Gamma^{(t)} +  \hat{\Sigma}_{x}^{1/2} \Pi^{(t+1)} \hat{\Sigma}_{x}^{1/2}  - H^{(t+1)}  $.

\end{enumerate}

\end{enumerate}
\end{algorithm}

%%%%%%%%%%%%%%%%%%%%%%%%%%%%%%
%%%%%%%%%%%%%%%%%%%%%%%%%%%%%%
% Tuning Parameter Selection
%%%%%%%%%%%%%%%%%%%%%%%%%%%%%%
%%%%%%%%%%%%%%%%%%%%%%%%%%%%%%
\subsection{Tuning parameter selection}
\label{subsec:tuning}
Our proposed method (\ref{Eq:variational3}) involves two user-specified  tuning parameters: the dimension $K$ of the central subspace $\cV_{y\mid x}$ and a sparsity tuning parameter $\rho$.
\citet{zhu2006sliced} used the Bayesian information criterion to select $K$.  Several authors proposed to select $K$ using bootstrap procedures \citep{ye2003using,dong2010dimension,ma2012semiparametric}.  In addition, sequential testing procedures were developed for determining $K$ \citep{li1991sliced,bura2001estimating,cookni2005,ma2013review}.

Motivated by \citet{cook2008principal}, we propose a cross-validation approach to select the tuning parameters $K$ and $\rho$.
Let $\hat{\Pi}$ be the solution of (\ref{Eq:variational3}), and recall that $\mathrm{span}(\hat{\Pi})$ is an estimate of the central subspace $\cV_{y\mid x}$.  
Let $\hat{\pi}_1,\ldots,\hat{\pi}_K$ be the top $K$ eigenvectors of $\hat{\Pi}$. 
Given a new data point $x^*$, define
\[
\hat{R}(x^*) = (\hat{\pi}_1^\T x^*,\ldots,\hat{\pi}_K^\T x^*)^\T, \qquad \qquad
w_i (x^*) = \frac{\exp  \left\{ -\frac{1}{2} \| \hat{R}({x}^*) - \hat{R} (x_i)  \|_2^2  \right\}}{\sum_{i=1}^n\exp  \left\{ -\frac{1}{2} \| \hat{R}({x}^*) - \hat{R} (x_i)  \|_2^2  \right\}},
\]
where $\|a\|_2 = (\sum_{j=1}^d a_j^2 )^{1/2}$ for $a\in \RR^d$.
The conditional mean $E(y \mid x =x^*)$ can then be estimated as  
\begin{equation}
\label{Eq:conditional mean}
\hat{E} (y\mid x = x^*) = \sum_{i=1}^n w_i (x^*)   y_i. 
\end{equation}
Details on the derivation of (\ref{Eq:conditional mean}) are deferred to \S~\ref{sec:pfc}. 

We propose an $M$-fold cross-validation procedure to select the tuning parameters $K$ and $\rho$ based on (\ref{Eq:conditional mean}). 
We first partition the $n$ observations into $M$ sets, $C_1,\ldots,C_M$.  
For each set $C_m$, we obtain an estimate of $\hat{\Pi}$ using all  observations outside the set $C_m$. We then predict the conditional mean for observations in $C_m$ using (\ref{Eq:conditional mean}).   The tuning parameters $K$ and $\rho$ are now chosen to minimize the overall prediction error$\sum_{m=1}^{M}\sum_{i \in C_m} \{y_i - \hat{E}(y \mid x = x_i)\}^2/(M|C_m|),$
where $|C_m|$ is the cardinality of the set $C_m$.

%%%%%%%%%%%%%%%%%%%%
% Estimation error
%%%%%%%%%%%%%%%%%%%%
\section{Theoretical Results}
\label{sec:theory}
We study the theoretical properties of the proposed estimator $\hat{\Pi}$ obtained from solving (\ref{Eq:variational3}) under the non-asymptotic setting in which $n$, $d$, $s$, and $K$ are allowed to grow. 
Throughout this section, we assume that the linearity condition in \eqref{Eq:linearity} holds and that $x_1,\ldots,x_n$ are independent random variables that are sub-Gaussian with covariance matrix $\Sigma_x$.  
Moreover, for simplicity, we assume that the largest generalized eigenvalue $\lambda_1$ is bounded by some constant, and that $K <\min (s,\log d)$.
%The elliptical distribution is needed for the linearity condition in \eqref{Eq:linearity} to hold. 
To quantify the distance between the estimated and population subspaces, we first establish a concentration result for 
$\Sigma_{E(x \mid y)}$ under the max norm.
Recall that $y_{(1)},\ldots,y_{(n)}$ are the order statistics of $y_1,\ldots,y_n$. 
Let $m\{y_{(i)}\} = E\{x \mid y_{(i)}\}$.  
We  state an assumption on the smoothness of  $m(y)$. 
\begin{assumption}
\label{ass:smoothness}
Let $B>0$ and let $\Xi_n (B)$ be the collection of all the $n$-point partitions $-B \le y_{(1)} \le \cdots \le y_{(n)}\le B$ on the interval $[-B,B]$. A vector-valued $m(y)$ is said to have a total variation of order $1/4$ if for any fixed $B>0$, 
\[
\underset{n\rightarrow \infty}{\lim}\;  \frac{1}{n^{1/4}} \underset{ \Xi_n (B)}{\sup}  \;\sum_{i=2}^{n} \|m\{y_{(i)}\}-m\{y_{(i-1)}\}\|_{\infty}=0,
\]
where $\| a\|_{\infty} = \max_{j} |a_j|$ for $a\in \RR^{d}.$ 
\end{assumption}
 A similar assumption is given by \citet{hsing1992asymptotic} and \citet{zhu1995asymptotics}, except that they considered the Euclidean norm on the quantity $m\{y_{(i)}\} - m\{y_{(i-1)}\}$ rather than the $\ell_\infty$ norm. 
In our problem, it suffices to assume the smoothness condition under the $\ell_\infty$ norm, since we are bounding the estimation error of $\hat{T}$ under the max norm. 
The following lemma provides an upper bound on the estimation error of $\hat{T}$ in (\ref{Eq:estimator conditional expectation}).
\begin{lemma}
\label{lemma:concentration conditional}
Assume that  $y_1,\ldots,y_n \in [-B,B]$ has a bounded support for some fixed $B>0$.    Assume that $x_1,\ldots,x_n$ are independent sub-Gaussian random variables with covariance matrix $\Sigma_x$. Under Assumption~\ref{ass:smoothness}, for sufficiently large $n$, there exists constants $C,C'>0$ such that with probability at least $1-\exp ( -C' \log d)$, 
\[
\|\hat{T}-T\|_{\max} =C   (\log d/n)^{1/2},
\]
where $\|A\|_{\max} = \max_{i,j} |A_{ij}|$ for $A \in \RR^{d\times d}$.
\end{lemma}
 For simplicity, we assume that $y$ has a bounded support in Lemma~\ref{lemma:concentration conditional}.
When $y$ is unbounded, a more refined analysis is needed to obtain an upper bound on the estimation error under additional assumptions on the inverse regression curve and the empirical distribution of $y$ \citep{zhu2006sliced}.
 Similar results can be shown for the estimator $\tilde{T}$ in (\ref{Eq:estimator conditional expectation2}).
We next state a result on the sample covariance matrix  $\hat{\Sigma}_{x}$, which follows from Lemma 1 of \citet{ravikumar2011high}.

\begin{proposition}
\label{proposition:covariance}
 Assume that $x_1,\ldots,x_n$ are independent sub-Gaussian random variables with the  covariance matrix $\Sigma_x$.  Let $\hat{\Sigma}_{x}$ be the sample covariance matrix.  Then there exists constants $C_1,C_1'>0$ such that 
\[
\|\hat{\Sigma}_{x} - \Sigma_{x}\|_{\max} = C_1 ( \log d/n)^{1/2}
\] 
with probability at least $1-\exp (-C_1' \log d)$.
\end{proposition}

\begin{corollary}
\label{cor:concentration}
Let $\hat{\Sigma}_{E(x\mid  y)} =  \hat{\Sigma}_x-\hat{T} $. 
Under the conditions in Lemma~\ref{lemma:concentration conditional} and Proposition~\ref{proposition:covariance}, there exists constants $C_2,C_2'>0$ such that
\[
\|\hat{\Sigma}_{E(x\mid y)}-\Sigma_{E(x \mid y)}\|_{\max} \le C_2 ( \log d/n)^{1/2}
\]
with probability at least $1-\exp (C_2' \log d)$.
\end{corollary}

Corollary~\ref{cor:concentration}  follows directly from Lemma~\ref{lemma:concentration conditional} and Proposition~\ref{proposition:covariance}.
Next, we state an assumption on  the $s$-sparse eigenvalue of $\Sigma_{x}$.  The assumption is commonly used in the high-dimensional literature  (see, for instance, \citealp{meinshausen2009lasso}). %The sparsity level $s$ is allowed to increase when $n$ and $d$ are increased.
\begin{assumption}
\label{ass:bounded eigenvalues}
The $s$-sparse minimal and  maximal eigenvalues of $\Sigma_{x}$ are
\begin{equation}
\label{Eq:sparse eigenvalues}
\lambda_{\min} (\Sigma_{x},s)  = \underset{v: \|v\|_{0}\le s }{\min} \frac{v^\T \Sigma_{x}v}{v^{\T}v}, \qquad  \qquad \lambda_{\max} (\Sigma_{x},s)  = \underset{v: \|v\|_{0}\le s }{\max} \frac{v^\T \Sigma_{x}v}{v^\T v},
\end{equation}
where $\|v\|_{0}$ is the number of non-zero elements in $v$.
Assume that there exists a constant $c>0$ such that 
$
c^{-1} \le \lambda_{\min}(\Sigma_{x},s) \le \lambda_{\max} (\Sigma_{x},s) \le c.
$
\end{assumption}

We now quantify the distance between the estimated and population subspaces.   
To this end, we establish the notion of distance between subspaces \citep{vu2013fantope}.  %Let $\|A\|_\mathrm{F} = (\sum_{i=1}^{d} \sum_{j=1}^d A_{ij}^2)^{1/2}$ be the Frobenius norm for $A\in \RR^{d\times d}$.
\begin{definition}
\label{def:subspace distance}
Let $\cV$ and $\hat{\cV}$ be $K$-dimensional subspaces of $\RR^{d}$.
Let $P_{\Pi}$ and $P_{\hat{\Pi}}$ be the projection matrices onto the subspaces $\cV$ and $\hat{\cV}$, respectively.  The distance between the two subspaces are defined as $D\bigl(\cV, \hat{\cV}\bigr) = \|P_{\Pi}-P_{\hat{\Pi}}  \|_\mathrm{F}$.
\end{definition}

The following theorem provides an upper bound on the subspace distance as defined in Definition~\ref{def:subspace distance} between $\Pi$ and the  solution $\hat{\Pi}$ obtained from solving~(\ref{Eq:variational3}). 
%%%%%%%%%%%%%%%%%%%%%%%%%%%%%%
% Theorem
%%%%%%%%%%%%%%%%%%%%%%%%%%%%%%
\begin{theorem} 
\label{theorem:upper bound}
Let $\cV$ and $\hat{\cV}$ be the true and estimated subspaces, respectively.  
Let $n > C s^2\log d/\lambda_K^2$ for some sufficiently large constant $C$, where $\lambda_K$ is the $K$th generalized eigenvalue of the pair of matrices $\{\Sigma_{E(x\mid y)},\Sigma_{x}  \}$.  Assume that $\lambda_K K^2 < s \log d$.
Let $\rho \ge C_1 (\log d/n)^{1/2}$ for some constant $C_1$.  
Under conditions in Corollary~\ref{cor:concentration} and Assumption~\ref{ass:bounded eigenvalues}, 
\[
D(\cV,\hat{\cV}) \leq C_2  s(\log d/n)^{1/2}/\lambda_K
\]
with probability at least $1-\exp (-C_3 s) - \exp ( -C_4 \log d)$ for some constants $C_2$, $C_3$, and  $C_4$.
\end{theorem}
Theorem~\ref{theorem:upper bound} states that with probability tending to one, the distance between the estimated and population subspaces is proportional to $s (\log d/n)^{1/2}/\lambda_K$ and decays to zero if $s=o\{\lambda_K(n / \log d)^{1/2}\}$.
That is, the number of active covariates cannot be too large. We will illustrate the results in Theorem~\ref{theorem:upper bound} in \S~\ref{sec:sim}. 

\begin{remark}
Our results allow the dimension $K$ to increase as a function of $n,d,s$ under the constraint that $\lambda_K = \omega \{s(\log d /n)^{1/2}\}$, where the notation $f(n)=\omega\{g(n)\}$ indicates $\lim_{n\rightarrow \infty} |f(n)/g(n)| \rightarrow \infty$.  In other words, the signal to noise ratio in terms of the $K$th generalized eigenvalue $\lambda_K$ has to be sufficiently large to attain a small estimation error.  We require that $\lambda_K K^2 < s \log d$, so $K$ cannot be too large compared to the number of active covariates.
\end{remark}

%%%%%%%%%%%%%%%%%%%%%%%%%%%%%%%%
%%%%%%%%%%%%%%%%%%%%%%%%%%%%%%%%
%%%%%%%%%%%%%%%%%%%%%%%%%%%%%%%%
%%%%%%%%%%%%%%%%%%%%%%%%%%%%%%%%
%%%%%%%%%%%%%%%%%%%%%%%%%%%%%%%%
%%%%%%%%%%%%%%%%%%%%%%%%%%%%%%%%
%%%%%%%%%%%%%%%%%%%%%%%%%%%%%%%%

%%%%%%%%%%%%%%%%%%%%%%%%%%%%%%
%%%%%%%%%%%%%%%%%%%%%%%%%%%%%%
% Simulation Studies
%%%%%%%%%%%%%%%%%%%%%%%%%%%%%%
%%%%%%%%%%%%%%%%%%%%%%%%%%%%%%
\section{Numerical Studies} 
\label{sec:sim}
We compare our proposal to three other methods on high-dimensional sparse sliced inverse regression under various simulation settings:  \citet{hilafu}, \citet{li2008sliced}, and \citet{wang2015estimating}.   
Recall from Definition~\ref{def:subspace sparsity} that subspace sparsity is determined by the diagonal elements of $\Pi$. 
Let $\hat{\Pi}$ be an estimator of $\Pi$.   We define the true positive rate as the proportion of correctly identified non-zero diagonals, and the false positive rate as the proportion of zero diagonals that are incorrectly identified to be non-zeros.  
Furthermore, we calculate the absolute correlation coefficient between the true sufficient predictor and its estimate.  
For simulation settings with $K>1$, we calculate the pairwise correlation between the estimated directions and each of the true sufficient dimension reduction directions.  We then select the maximum pairwise correlation for each of the true direction and take their average.
In addition, we compute the subspace distance between the true and estimated subspace to illustrate the theoretical result in Theorem~\ref{theorem:upper bound}.

We simulated $x$ from $N_d (0,\Sigma_x)$, where $(\Sigma_x)_{ij} = 0.5^{|i-j|}$ for $1\le i,j \le d$,  $\epsilon$ from $N(0,1)$, and employed the following regression models:
\begin{enumerate}
\item A linear regression model with three active predictors:
\[
y =(x_1+x_2+x_3)/3^{1/2} + 2 \epsilon.
\] 
In this setting, the central subspace is spanned by the directions $\beta = (1_3,0_{d-3})^\T$ and $K=1$.  

\item A non-linear regression model with three active predictors:  
\[
y=1+ \exp \{(x_1+x_2+x_3 )/3^{1/2}\} + \epsilon.
\]
  This regression model has recently been considered in \citet{hilafu}. In this study, the central subspace is spanned by the direction $\beta=(1_3, 0_{d-3})^\T$ and $K=1$.

\item A non-linear regression model with five active predictors:
\[
y = \frac{x_1+x_2+x_3}{0.5+(x_4+x_5+1.5)^2} + 0.1 \epsilon.
\] 
This simulation setting is similar to that of \citet{chen2010coordinate}.  In this study, the central subspace is spanned by the directions $\beta_1 = (1_3,{0}_{d-3})^\T$, $\beta_2 = ({0}_3,{1}_2,{0}_{d-5})^\T$, and $K=2$.

\end{enumerate}

Sliced inverse regression requires estimators of the marginal and conditional covariance matrices, $\Sigma_{x}$ and $\Sigma_{E(x \mid y)}$.
We estimated $\Sigma_{x}$ using the sample covariance matrix  $\hat{\Sigma}_{x}$.  Then, $\Sigma_{E(x \mid y)}$ can be estimated using the identity  $\hat{\Sigma}_{E(x \mid y)} = \hat{\Sigma}_{x} - \tilde{T}$, where $\tilde{T}$ is defined in (\ref{Eq:estimator conditional expectation2}).  We constructed $\tilde{T}$ with $H=5$ slices.
There are two tuning parameters in our proposal (\ref{Eq:variational3}), which we selected using the cross-validation idea outlined in \S~\ref{subsec:tuning}. 
Similarly, we used cross-validation to select tuning parameters for \citet{wang2015estimating}.
For the proposal in \citet{li2008sliced}, the authors proposed three different methods for selecting the tuning parameters: we performed tuning parameter selection with these three methods and reported only the best results for \citet{li2008sliced}.
We considered multiple set of tuning parameters for \citet{hilafu} and reported only the best results for their proposal. 
The true and false positive rates, and the absolute correlation coefficient, averaged over 200 data sets, are reported in Table~\ref{table1}.

  %%%%%%%%%%%%%%%%%%%%%%%%%%%%
% Table 1
%%%%%%%%%%%%%%%%%%%%%%%%%%%%
\begin{table}[!htp]
\footnotesize
\begin{center}
\caption{True and false positive rates, and absolute correlation coefficient with $n=(100, 200)$ and $d=150$.   The mean (standard error), averaged over 200 data sets, are reported.  All entries are multiplied by 100. TPR, true positive rate; FPR, false positive rate; corr, absolute correlation coefficient.  }
\begin{tabular}{cl    c cc cccc}
  \hline
& & \multicolumn{3}{c}{$n=100$ and  $d=150$}& \multicolumn{3}{c}{$n=200$ and  $d=150$} \\
&&Setting 1&Setting 2&Setting 3&Setting 1&Setting 2&Setting 3& \\ \hline
&TPR & 96 (1) &  94$\cdot$2  (1$\cdot$2)  & 91$\cdot$3 (1$\cdot$1) & 98$\cdot$2 (0$\cdot$5)  &  98$\cdot$5 (0$\cdot$5) &98$\cdot$9 (2$\cdot$5)\\
Our proposed method &FPR & 6 (0$\cdot$9) & 3$\cdot$6 (0$\cdot$7) &  7$\cdot$4 (0$\cdot$1)    & 3$\cdot$4 (0$\cdot$4)  &  1$\cdot$1 (0$\cdot$2) &2$\cdot$5 (0$\cdot$3)\\ 
&corr& 88$\cdot$3 (0$\cdot$9) &  86$\cdot$4 (1$\cdot$1)  &74$\cdot$2 (1$\cdot$1)  & 90$\cdot$9 (0$\cdot$5)  & 92$\cdot$1 (0$\cdot$5) & 79$\cdot$2 (0$\cdot$6)\\ \hline
&TPR& 95$\cdot$3 (0$\cdot$9) &  100 (0)  & 99$\cdot$6 (0$\cdot$4) & 100 (0)  & 100 (0) &100 (0)\\
\citet{hilafu}&FPR& 4$\cdot$9 (0$\cdot$1) &  4$\cdot$8 (0$\cdot$1)  &3$\cdot$5 (0$\cdot$1)& 5$\cdot$9 (0$\cdot$2) & 6$\cdot$7 (0$\cdot$3)& 4$\cdot$5 (0$\cdot$2)\\ 
&corr& 59$\cdot$2 (1$\cdot$1) &87$\cdot$8 (0$\cdot$5) &78$\cdot$8 (0$\cdot$6)  & 78 (0$\cdot$6)  & 94$\cdot$2 (0$\cdot$2)& 87$\cdot$4 (0$\cdot$5)\\ \hline
&TPR& 97$\cdot$8 (0$\cdot$1) &  98$\cdot$1 (0$\cdot$1)  & 97$\cdot$8 (0$\cdot$1) & 98$\cdot$9 (0$\cdot$1)   & 99$\cdot$1 (0$\cdot$1)  &97$\cdot$9 (0$\cdot$1) \\
\citet{li2008sliced}&FPR& 8$\cdot$3 (1$\cdot$2) &  3$\cdot$8 (0$\cdot$8)  &23$\cdot$4 (1$\cdot$1)& 1$\cdot$2 (0$\cdot$4) & 0$\cdot$3 (0$\cdot$2)& 19$\cdot$7 (1$\cdot$1)\\ 
&corr& 84$\cdot$3 (0$\cdot$9) &88$\cdot$9 (0$\cdot$6) &62$\cdot$7 (0$\cdot$7)  & 93$\cdot$6 (0$\cdot$4)  & 95$\cdot$8 (0$\cdot$3)& 69$\cdot$7 (0$\cdot$5)\\ \hline
&TPR& 88$\cdot$8 (1$\cdot$5) &  93$\cdot$5 (1$\cdot$2)  & 80$\cdot$1 (1$\cdot$2) & 97$\cdot$5 (1$\cdot$0)   & 98$\cdot$8 (0$\cdot$7)  &96$\cdot$3 (0$\cdot$6) \\
\citet{wang2015estimating}&FPR& 0$\cdot$6 (0$\cdot$1) &  0$\cdot$6 (0$\cdot$1)  &0$\cdot$2 (0$\cdot$1)& 0$\cdot$3 (0$\cdot$1) & 0$\cdot$3 (0$\cdot$1)& 0$\cdot$1 (0$\cdot$1)\\ 
&corr& 81$\cdot$5 (1$\cdot$4) &85$\cdot$1 (1$\cdot$3) &69$\cdot$9 (1$\cdot$1)  & 91$\cdot$3 (1$\cdot$1)  & 93$\cdot$2 (1$\cdot$0)& 84$\cdot$4 (0$\cdot$7)\\ \hline
\end{tabular}
\label{table1}
\end{center}
\end{table}

Table~\ref{table1} shows that the proposed method performs competitively against recent proposals for high-dimensional sliced inverse regression \citep{hilafu,wang2015estimating,li2008sliced}.  %In particular, our approach has zero false positive rate across all simulation settings, indicating that it is more conservative than all three other methods in terms of selecting the active covariates.  
In the low-dimensional setting when $n=200$, our method performs competitively with all of the existing methods across all three settings.  
In the high-dimensional setting when $n=100$, for setting one, our proposal yields the best absolute correlation between the true and estimated sufficient dimension direction.
%Our results are comparable to that of \citet{li2008sliced} and \citet{wang2015estimating}.  
All methods perform similarly in setting two. 
Setting three is a harder problem and the method of \citet{li2008sliced} has an extremely high false positive rate. 
The method of \citet{wang2015estimating} has the lowest 
true positive rate and a low correlation, and that of    
\citet{hilafu} slightly outperforms our proposal in terms of true positive rate and correlation.  
However, the tuning parameters for our proposals are selected entirely using cross-validation and that we report the best results for \citet{hilafu} after considering multiple tuning parameters.
Moreover, \citet{hilafu} has the worst performance in setting one. 
In short, our proposed method is the most robust proposal across all three settings in the high-dimensional setting.

\begin{figure}[!t]
\begin{center}
   \subfigure[]{\includegraphics[scale=0.365]{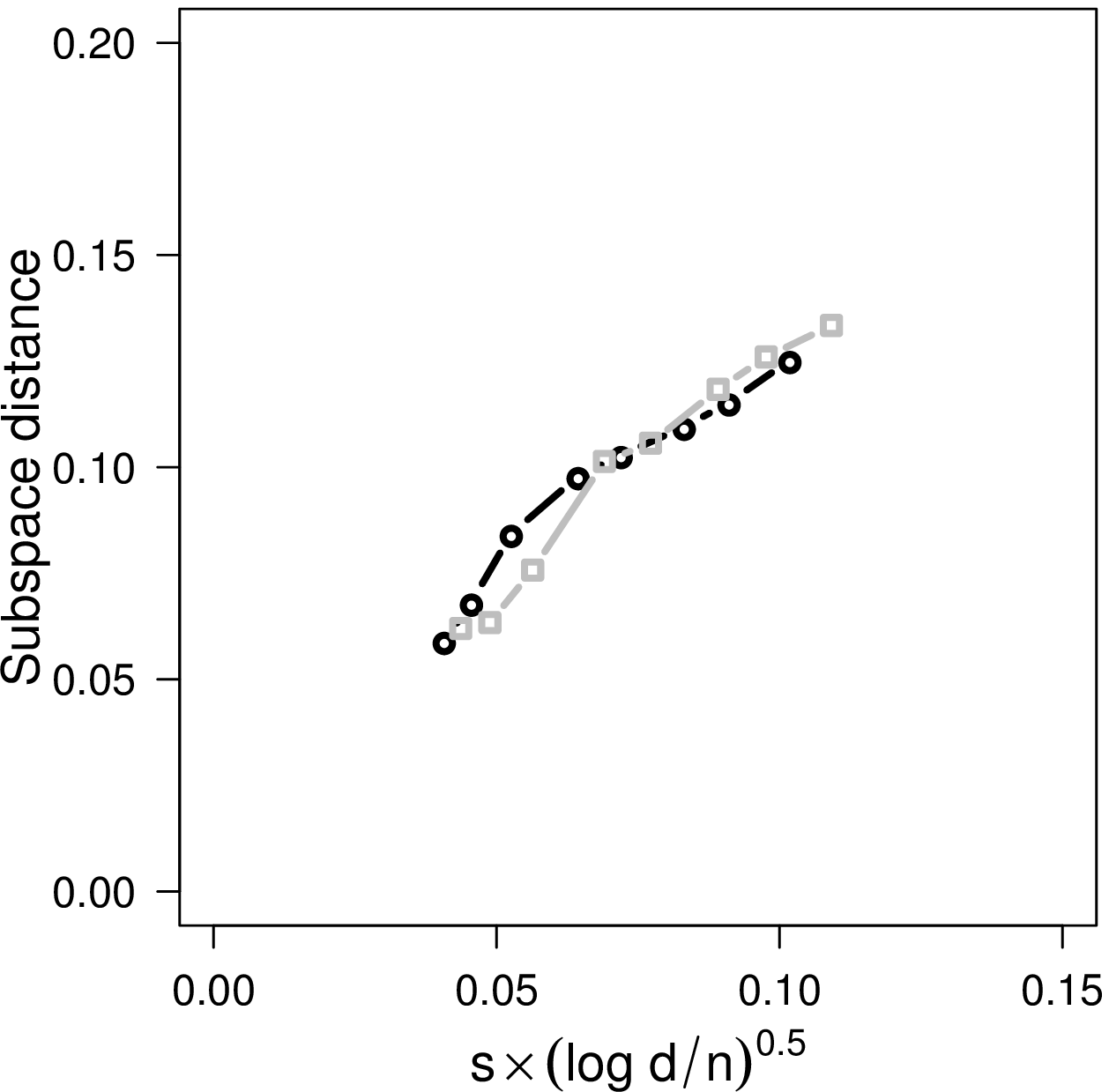}}\;
   \subfigure[]{\includegraphics[scale=0.365]{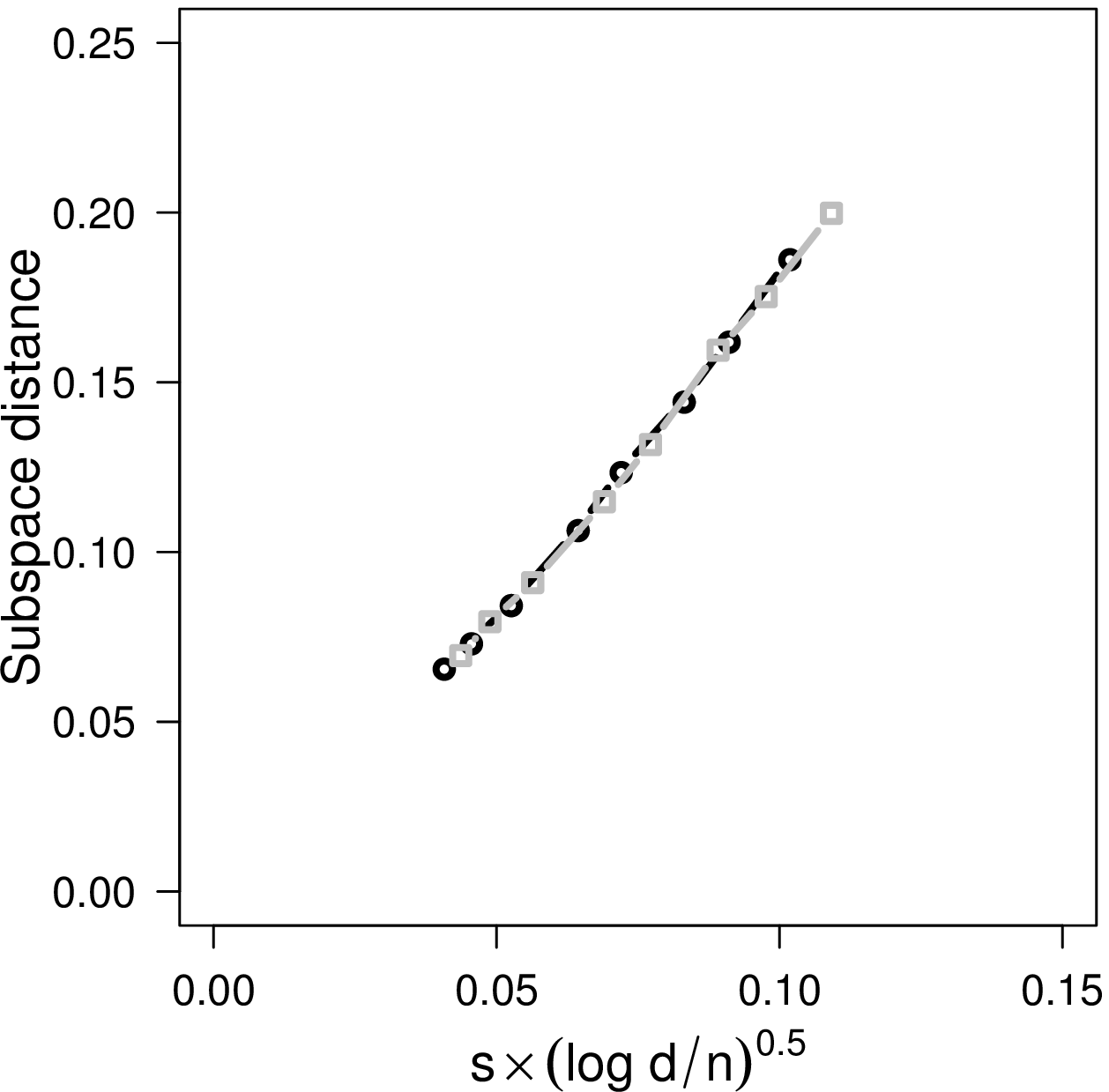}}\;
   \subfigure[]{\includegraphics[scale=0.365]{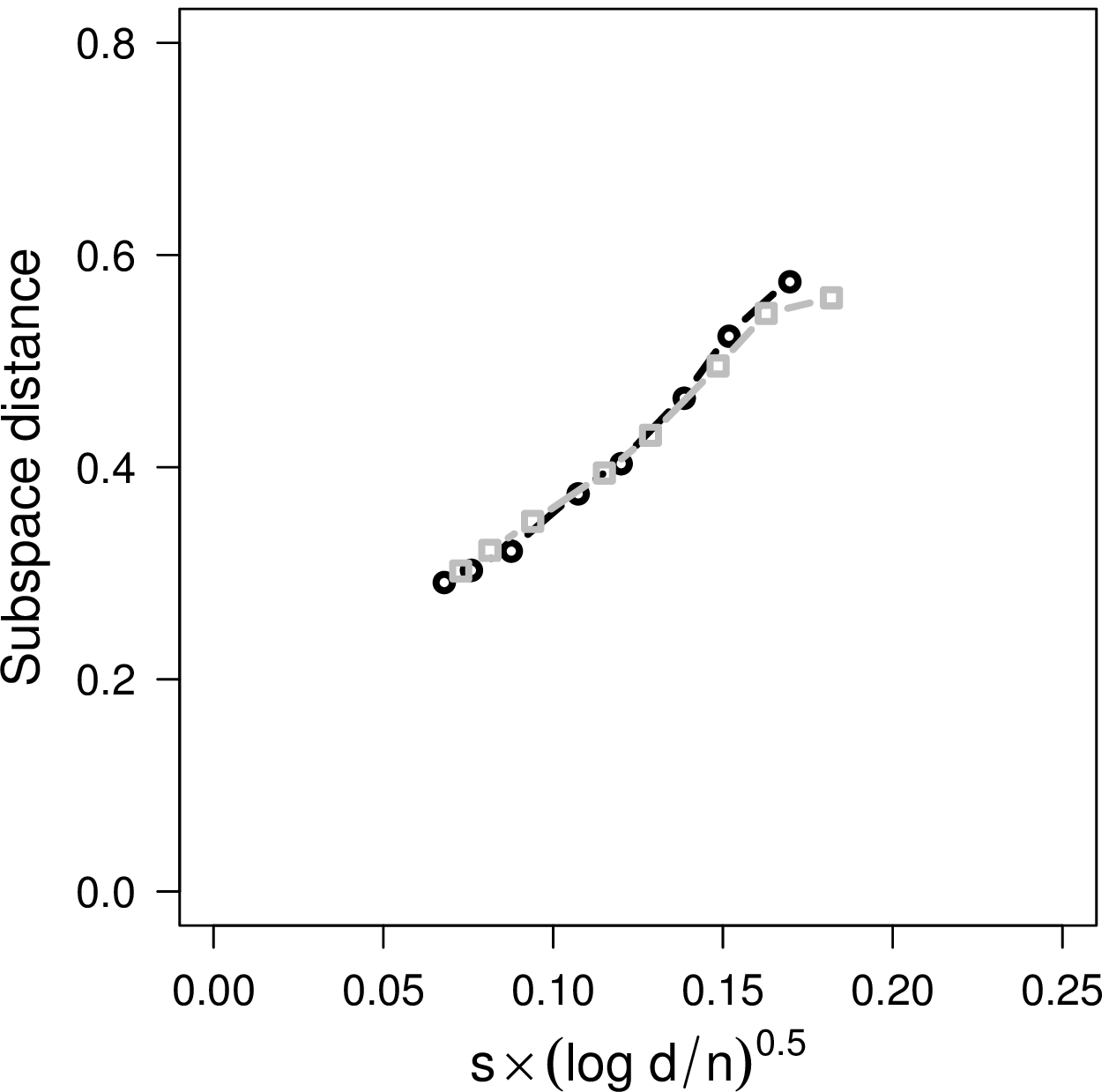}}
\end{center}
\caption{\label{fig:sinfracsim}  Results for the subspace distance, averaged over 500 data sets.  Panels (a), (b), and (c) are the results for simulation settings 1, 2, and 3, respectively. 
The lines are obtained by varying the sample size $n$ with $d=100$ (circle black line) and $d=200$ (square gray line), respectively.}
\end{figure}

Next, we evaluated the distance between the estimated and the population subspaces.  We assume that $K$ is known, and select $\rho = 2  (\log d/n)^{1/2}$ as suggested by  Theorem~\ref{theorem:upper bound}. 
The results for $d=(100,200)$ as a function of $n$, averaged over 500 data sets, are presented in Figures~\ref{fig:sinfracsim}(a)--(c).  
The subspace distance between the estimated and population subspaces is indeed proportional to $s(\log d/n)^{1/2}$.

% ---------------------------------------------------------------------------%
% ---------------------------------------------------------------------------%
%%%%%%%%%%%%%%%%%%%%%%%%%%%%%%
%%%%%%%%%%%%%%%%%%%%%%%%%%%%%%
% Principal Fitted Components
%%%%%%%%%%%%%%%%%%%%%%%%%%%%%%
%%%%%%%%%%%%%%%%%%%%%%%%%%%%%%
% ---------------------------------------------------------------------------%
% ---------------------------------------------------------------------------%
\section{An Extension to Sparse Principal Fitted Components}
\label{sec:pfc}
We briefly outline an extension of the proposed method for principal fitted components in the high-dimensional setting. 
\citet{cook2008principal} proposed several model-based sufficient dimension reduction methods, collectively referred to as principal fitted components. 
Let $x_y$ be the conditional random variable of $x$ given $y$. Assume that $x_y$ is normally distributed from $N_d(\mu_y,\Delta)$. Furthermore, let $\bar{\mu} = E(x)$, and let $\cV_{\Gamma} = \mathrm{span} (\mu_y -\bar{\mu} \mid y\in \mathcal{S}_y)$, where $\Gamma \in \RR^{d\times K}$ denotes a semi-orthogonal matrix whose columns form a basis for the $K$-dimensional subspace $\cV_{\Gamma}$, and $\mathcal{S}_y$ denotes the sample space of $y$.
  \citet{cook2008principal}
considered the inverse regression model
\begin{equation}
\label{PFC}
x=\bar{\mu} + \Gamma \xi \{f(y) -\bar{f}(y)\} + \Delta^{1/2} \epsilon,
\end{equation}
where $\xi \in \RR^{K\times r}$ is an unrestricted rank $K$ matrix with  $K < r$, $f(y) \in  \RR^r$  is a known vector-valued function of $y$, and $\epsilon$ is $N(0,I_d)$ that is independent of $y$. The covariates $f(y)$ usually takes the form of polynomial, piecewise linear, or Fourier basis functions.  Thus, the regression model~(\ref{PFC}) can effectively model nonlinear relationships between the covariates and the response. 
Principal fitted components  yields sliced inverse regression as a special case when $y$ is categorical \citep{cook2008principal}.

Under model~(\ref{PFC}), \citet{cook2008principal} showed that the maximum likelihood estimator of the central subspace $\cV_{\Gamma}$ can be obtained by solving the generalized eigenvalue problem $\hat{\Sigma}_{\mathrm{fit}}  V =  \hat{\Sigma}_x V \Lambda$,
where $\hat{\Sigma}_{\mathrm{fit}}$ is the sample covariance matrix of the estimated vectors from the linear regression of $x$ on $f$.
More specifically, let $\mathbb{X}$ denote the $n\times d$ matrix with rows $(x-\bar{x})^\T$ and let $\mathbb{F}$ denote the $n\times r$ matrix with rows
$\{f(y)-\bar{f}(y)\}^\T$.
Then, $\hat{\Sigma}_{\mathrm{fit}} = \mathbb{X}^\T \mathbb{F}(\mathbb{F}^\T\mathbb{F})^{-1}\mathbb{F}^\T \mathbb{X}/n$ and $\hat{\Sigma}_x = \mathbb{X}^\T \mathbb{X}/n$.  
While the estimator of the central subspace is derived under the normality assumption, it is also robust to non-normal error (\citealp{cook2008principal}, Theorem 3$\cdot$5).  Therefore, normality assumption on the covariates is not crucial to the principal fitted components.

A convex relaxation for the principal fitted components takes the form
 \begin{equation}
 \label{Eq:pfc convex}
\underset{\Pi\in \mathcal{M}}{\mathrm{minimize}} \; -\mathrm{tr}\left(\hat{\Sigma}_{\mathrm{fit}} \Pi\right) + \rho  \|\Pi\|_{1} \; \text{subject~to~} \|\hat{\Sigma}_{x}^{1/2} \Pi \hat{\Sigma}_{x}^{1/2}\|_{*}\le K, \; \|\hat{\Sigma}_{x}^{1/2} \Pi \hat{\Sigma}_{x}^{1/2}\|_{\mathrm{sp}}\le 1.
 \end{equation}
Algorithm~\ref{Alg:ssliced inverse regression} can directly be adapted to solve (\ref{Eq:pfc convex});
with some abuse of notation, let $\hat{\Pi}$ be the solution to (\ref{Eq:pfc convex}) and let $\hat{\pi}_1,\ldots,\hat{\pi}_K$ be the $K$ largest eigenvector of $\hat{\Pi}$.

One of the main advantages of principal fitted components is that a model for $x$ given $y$ can be inverted to provide a method for estimating the mean function $E(y\mid x)$ without specifying a model for the joint distribution $(y,x)$. Let $R(x)$ be the $K$-dimensional sufficient reduction.   Let $g(x\mid y)$ and $g\{R(x) \mid y\}$ be the conditional densities of $x$ given $y$ and $R(x)$ given $y$. Then, the conditional expectation can be written as 
\[
E(y \mid x)  = E\{y \mid R(x)\}       = \frac{E [y  g\{R(x) \mid y\}] }{E[g\{R(x)\mid y\}]},
\]
where the expectation is taken with respect to the random variable $y$.
Under the normality assumption on $x_y$, for a new data point $x^*$, the conditional mean  can be estimated as 
\[
\hat{E} (y \mid x = x^*)  = \sum_{i=1}^n w_i (x^*) y_i, \;\; \;\;
w_i (x^*) = \frac{\exp  \left\{ -\frac{1}{2} \| \hat{R}(x^*) - \hat{R} (x_i)  \|_2^2  \right\}}{\sum_{i=1}^n\exp  \left\{ -\frac{1}{2} \| \hat{R}(x^*) - \hat{R} (x_i)  \|_2^2  \right\}},
\]
where $\hat{R}(x^*) = (\hat{\pi}_1^\T x^*,\ldots,\hat{\pi}_K^\T x^*)^\T$ is an estimate of the $K$-dimensional sufficient reduction.
This motivates the cross-validation procedure described in \S~\ref{subsec:tuning} for selecting the tuning parameters $K$ and $\rho$.

%%%%%%%%%%%%%%%%%%%%%%%%%
%%%%%%%%%%%%%%%%%%%%%%%%%
% Discussion
%%%%%%%%%%%%%%%%%%%%%%%%%
%%%%%%%%%%%%%%%%%%%%%%%%%
\section{Discussion}
\label{sec:discussion}
We have proposed a convex relaxation for sparse sliced inverse regression in the high-dimensional setting, using the fact that sliced inverse regression is a special case of the generalized eigenvalue problem.
As discussed in \citet{chen2010coordinate} and \citet{li2007sparse}, many other sufficient dimension reduction methods 
can be formulated as sparse generalized eigenvalue problems.  These include sliced average variance estimation, directional regression, principal fitted components, principal hessian direction, and iterative hessian transformation.
Therefore, these models can all be applied using the proposed method in \eqref{Eq:variational3} with different choices of covariance matrices.

Many sufficient dimension reduction methods rely on the linearity condition (\ref{Eq:linearity}), but this  is not always satisfied. 
To address this, \citet{ma2012semiparametric} proposed a semiparametric approach for sufficient dimension reduction that removes the linearity condition.    In future work, it will be of interest to propose a high-dimensional semiparametric approach for sufficient dimension reduction using recently developed theoretical tools in high-dimensional statistics.

Many authors have proposed methods to estimate the subspace dimension $K$.
These include the Bayesian information criterion, the bootstrap, and sequential testing  \citep{zhu2006sliced,ye2003using,dong2010dimension,ma2012semiparametric,li1991sliced,bura2001estimating,cookni2005}.
\citet{ma2015validated} proposed a validated information criterion for selecting $K$ in dimension reduction models. 
However, these methods are not directly applicable to the high-dimensional setting.    
It will be of interest to develop a principled way to estimate the subspace dimension $K$ consistently in this setting.

%%%%%%%%%%%%%%%%%%%%%%%%%%%%%%%%%%%%
%%%%%%%%%%%%%%%%%%%%%%%%%%%%%%%%%%%%
% Derivation of Algorithm
%%%%%%%%%%%%%%%%%%%%%%%%%%%%%%%%%%%%
%%%%%%%%%%%%%%%%%%%%%%%%%%%%%%%%%%%%
\appendix
\section{Derivation of Algorithm~\ref{Alg:ssliced inverse regression}}
\label{appendix:alg}
 In this section, we derive the linearized alternating direction methods of multiplier algorithm for solving (\ref{Eq:variational3}).
Optimization problem (\ref{Eq:variational3}) is equivalent to 
\begin{equation}
\label{Eq:alg1}
\underset{\Pi,H\in \mathcal{M}}{\text{minimize}} \quad -\mathrm{tr}\left\{\hat{\Sigma}_{E(x \mid y)} \Pi\right\} + \rho  \|\Pi\|_{1} + g(H) \qquad \mathrm{subject \; to\;} \hat{\Sigma}_{x}^{1/2} \Pi \hat{\Sigma}_{x}^{1/2} = H,
\end{equation}
where $g(H) = \infty 1_{( \|H\|_* > K  )} +\infty 1_{( \|H\|_{\mathrm{sp}} > 1  )}  $.  
  The scaled augmented Lagrangian for (\ref{Eq:alg1}) takes the form
\begin{equation}
\label{Eq:alg2}
L(\Pi,H,\Gamma) = -\mathrm{tr}\left\{\hat{\Sigma}_{E(x \mid y)} \Pi\right\} + \rho \|\Pi\|_{1} + g(H) 
+ \frac{\nu}{2}   \left\| \hat{\Sigma}_{x}^{1/2} \Pi \hat{\Sigma}_{x}^{1/2} - H + \Gamma \right\|_\mathrm{F}^2.
\end{equation}
The proposed algorithm requires the updates for $\Pi$, $H$, and $\Gamma$.  We now proceed to derive the updates for $\Pi$ and $H$.\\

%%%%%%%%%%%%%%%%%%%%
% Update for Pi
%%%%%%%%%%%%%%%%%%%%
Update for $\Pi$: To obtain a closed-form update for $\Pi$, we linearize the quadratic term in the scaled augmented Lagrangian (\ref{Eq:alg2}) as suggested by  \citet{fang2015generalized}.
For two matrices $A,B \in \RR^{d\times d}$, we use the identity $\mathrm{vec}\left( A B A  \right) = \left(A^\T \otimes A\right) \mathrm{vec}(B)$, where $\mathrm{vec} (\cdot)$ is the vectorization operation which converts a matrix into a column vector and $\otimes$ is the Kronecker product. 
Let $\pi = \mathrm{vec}({\Pi})$, $h = \mathrm{vec}(H)$, and $\gamma = \mathrm{vec}(\Gamma)$.
Thus, an update for $\pi$ can be obtained by minimizing
\begin{equation}
\label{Eq:alg Pi1}
\begin{split}
- \mathrm{vec} \left\{\hat{\Sigma}_{E(x \mid y)}\right\}^\T \pi  + \rho \|\pi\|_1 + 
\frac{\nu}{2} \left\|\left( \hat{\Sigma}_{x}^{1/2}\otimes\hat{\Sigma}_{x}^{1/2}        \right) \pi -h+\gamma\right\|_2^2.
\end{split}
\end{equation}
 However, there is no closed-form solution for $\pi$ due to the quadratic term in~(\ref{Eq:alg Pi1}). 
 %We instead consider the linearized alternating direction method of multipliers algorithm to obtain a closed-form solution for updating $\pi$.  The convergence of linearized alternating direction method of multipliers algorithm algorithm has been studied in \citet{fang2015generalized}. 

Similar to that of \citet{fang2015generalized}, we 
linearize  the quadratic term in (\ref{Eq:alg Pi1})  by applying a second-order Taylor Expansion and obtain the following update for $\pi$:
\begin{equation}
\small
\label{Eq:alg Pi2}
\pi^{(t+1)} = \underset{\pi}{\mathrm{argmin}} \;  \left[ - \mathrm{vec} \left\{\hat{\Sigma}_{E(x \mid y)}\right\}^\T \pi  + \rho  \|\pi\|_1 + 
\nu \left\{\pi - \pi^{(t)}\right\}^\T m^{(t)}   + \frac{\tau}{2}\|\pi-\pi^{(t)}\|_2^2 \right],
\end{equation}
where $\pi^{(t)}$ is the value of $\pi$ at the $t$th iteration and 
\[
m^{(t)} =( \hat{\Sigma}_{x}^{1/2}\otimes\hat{\Sigma}_{x}^{1/2}  ) \{  \hat{\Sigma}_{x}^{1/2}\otimes\hat{\Sigma}_{x}^{1/2} \pi^{(t)} -h^{(t)}+\gamma^{(t)}\}.
\]
  As suggested by \citet{fang2015generalized}, we pick $\tau > 2\nu \lambda_{\max}^2 (\hat{\Sigma}_{x})$ to ensure the convergence of the linearized alternating direction method of multipliers algorithm. 
  
Problem (\ref{Eq:alg Pi2}) is equivalent to 
%%%%%%%%%%%%%%%%%%%%%%%%%%%%%%
\begin{equation}
\small
\label{Eq:alg Pi3}
\Pi^{(t+1)} = \underset{\Pi \in \mathcal{M}}{\mathrm{argmin}} \; \rho \|\Pi\|_{1} + \frac{\tau}{2}  \left\| \Pi - \left[\Pi^{(t)} + \frac{1}{\tau} \hat{\Sigma}_{E(x \mid y)} -\frac{\nu}{\tau} \hat{\Sigma}_{x} \Pi^{(t)} \hat{\Sigma}_{x} + \frac{\nu}{\tau} \hat{\Sigma}_{x}^{1/2} \{H^{(t)}-\Gamma^{(t)}\} \hat{\Sigma}_{x}^{1/2}        \right]    \right\|_\mathrm{F}^2,
\end{equation}
which has the closed-form solution 
\[
 \Pi^{(t+1)} = \mathrm{Soft}\left( \Pi^{(t)}+   \frac{1}{\tau} \hat{\Sigma}_{E(x\mid y)} -
\frac{\nu}{\tau}\left[\hat{\Sigma}_{x} \Pi^{(t)} \hat{\Sigma}_{x} -\hat{\Sigma}_{x}^{1/2}\left\{  H^{(t)}- \Gamma^{(t)}  \right\} \hat{\Sigma}_{x}^{1/2}\right],  \frac{\rho}{\tau}      \right).
\] 
Here, $\mathrm{Soft}$ denotes the soft-thresholding operator, applied element-wise to a matrix:  $\mathrm{Soft}(A_{ij},b) = \text{sign}(A_{ij})  \max\left( |A_{ij}|-b, 0\right)$.\\

%%%%%%%%%%%%%%%%%%%%
% Update for H
%%%%%%%%%%%%%%%%%%%%
Update for $H$: The update for $H$ can be obtained as 
\begin{equation}
\label{Eq:alg H1}
H^{(t+1)} = \underset{H \in \mathcal{M}}{\mathrm{argmin}} \;  g(H)+ \frac{\nu}{2} \left\| H -\left\{\hat{\Sigma}_{x}^{1/2} \Pi^{(t+1)} \hat{\Sigma}_{x}^{1/2} + \Gamma^{(t)} \right\} \right\|_\mathrm{F}^2,
\end{equation}
which has a closed-form solution.  The following proposition follows directly from Lemma 4.1 in \citet{vu2013fantope} and  Proposition 10.2 in \citet{gao2014sparse}.

%%%%%%%%%%%%%%%%%%%%%%%%%%%
% Proposition for updating H
%%%%%%%%%%%%%%%%%%%%%%%%%%%
\begin{proposition}
\label{prop:update for H}
Let $\sum_{j=1}^d \omega_j {u}_j {u}_j^\T$ be the singular value decomposition of $W$.  Let $H^*$ be the solution to the optimization problem
\[
\underset{H \in \mathcal{M}}{\mathrm{minimize}} \; \|H- W  \|_\mathrm{F} \qquad \mathrm{subject\; to\;} \|H\|_* \le K \mathrm{ \; and \; } \|H\|_\mathrm{sp} \le 1.
\]
Then, $H^* = \sum_{j=1}^d \min \{ 1, \max \left( \omega_j - \gamma^*,0\right) \} {u}_j {u}_j^\T$, where 
\[
\gamma^* = \underset{\gamma>0 }{\mathrm{argmin}}\; \gamma \qquad \mathrm{subject \; to\;} \sum_{j=1}^d \min \left\{ 1, \max \left(\omega_j-\gamma,0\right)\right\}\le K.
\]
\end{proposition}

 Let $\sum_{j=1}^d \omega_j {u}_j{u}_j^\T$ be the singular value decomposition of  $ \Gamma^{(t)} + \hat{\Sigma}_{x}^{1/2} \Pi^{(t+1)} \hat{\Sigma}_{x}^{1/2}$. Thus, by proposition (\ref{prop:update for H}), we have 
\[
H^{(t+1)} = \sum_{j=1}^d \min \left\{ 1, \max \left(\omega_j-\gamma^*,0\right)\right\} {u}_j {u}_j^\T,
\] 
where
\begin{equation}
\label{Eq:alg H2}
\gamma^* = \underset{\gamma>0 }{\mathrm{argmin}}\; \gamma \qquad \mathrm{subject \; to\;} \sum_{j=1}^d \min \left\{ 1, \max \left(\omega_j-\gamma,0\right)\right\}\le K.
\end{equation}

%%%%%%%%%%%%%%%%%%%%%%%%%%%
% Update for Gamma
%%%%%%%%%%%%%%%%%%%%%%%%%%%
Finally, the update for the dual variable $\Gamma$ takes the form 
\[
\Gamma^{(t+1)}  = \Gamma^{(t)} + \hat{\Sigma}_{x}^{1/2} \Pi^{(t+1)} \hat{\Sigma}_{x}^{1/2}-H^{(t+1)}.
\]

%%%%%%%%%%%%%%%%%%%%
% Proof of Concentration
%%%%%%%%%%%%%%%%%%%%
\section{Proof of Lemma~\ref{lemma:concentration conditional}}
\label{appendix:concentration}
Recall from (\ref{Eq:estimator conditional expectation}) that
\begin{equation}
\label{Eq:concentration conditional1}
\hat{T} = \frac{1}{n} \sum_{i=1}^{\lfloor n/2\rfloor} \{x_{(2i)^*}-x_{(2i-1)^*}\} \{x_{(2i)^*}-x_{(2i-1)^*}\}^\T.
\end{equation}
In this section, we show that the estimation error between $\hat{T}$ and $T$ is bounded above under the max norm. Recall that $y_{(i)}$ is the $i$th order statistic of $y_1,\ldots,y_n$ and $x_{(i)^*}$ is the value of $x$ corresponding to $y_{(i)}$.
Denote the inverse regression curve and its residual by
\[
m(y) = E(x \mid y) \qquad \mathrm{and}\qquad \epsilon = x -m(y).
\]
It is shown in \citet{yang1977general} that the concomitants $\epsilon_{(i)^*} = x_{(i)^*} - m \{y_{(i)}\}$ are conditionally independent with mean zero, given the order statistics $y_{(i)}$.  We denote the $j$th element of $m(y)$, $\epsilon$, and $x$ as $m_j (y)$, $\epsilon_j$, and $x_j$, respectively.
To prove Lemma~\ref{lemma:concentration conditional}, we state some properties of the residual in the following proposition.
The proof of the following proposition is a direct consequence of Jensen's inequality and the properties of sub-Gaussian random variables.
\begin{proposition}
\label{prop:property}
 Assume that $x$ is sub-Gaussian with mean zero and  covariance matrix $\Sigma_x$.  Then, $x_j$ has a sub-Gaussian norm $\|x_j\|_{\psi_2} \le C \{(\Sigma_{x})_{jj}\}^{1/2} $, where $\|x_j\|_{\psi_2} = \sup_{p\ge 1} \; p^{-1/2} (E |x_j|^p)^{1/p}$.  Moreover, $m_j(y)$ and $\epsilon_j$ are sub-Gaussian with sub-Gaussian norm $\|x_j\|_{\psi_2}$ and $2\|x_j\|_{\psi_2} $, respectively.
\end{proposition}

We now prove Lemma~\ref{lemma:concentration conditional}.  Substituting $x = m(y)  + \epsilon$ into (\ref{Eq:concentration conditional1}),
we have
\begin{equation}
\label{Eq:concentration conditional2}
\begin{split}
\hat{T} &= \frac{1}{n} \sum_{i=1}^{\lfloor n/2 \rfloor}\{x_{(2i)^*}-x_{(2i-1)^*}\}\{x_{(2i)^*}-x_{(2i-1)^*}\}^\T\\
&= \frac{1}{n} \sum_{i=1}^{\lfloor n/2 \rfloor} [m\{y_{(2i)}\}-m\{y_{(2i-1)}\}][m\{y_{(2i)}\}-m\{y_{(2i-1)}\}]^\T+ \frac{1}{n} \sum_{i=1}^{\lfloor n/2 \rfloor} \{\epsilon_{(2i)^*}-\epsilon_{(2i-1)^*}\} [m\{y_{(2i)}\}-m\{y_{(2i-1)}\}]^\T\\
&+\frac{1}{n} \sum_{i=1}^{\lfloor n/2 \rfloor}  [m\{y_{(2i)}\}-m\{y_{(2i-1)}\}] \{\epsilon_{(2i)^*}-\epsilon_{(2i-1)^*}\}^\T + \frac{1}{n} \sum_{i=1}^{\lfloor n/2 \rfloor}  \{\epsilon_{(2i)^*}-\epsilon_{(2i-1)^*}\} \{\epsilon_{(2i)^*}-\epsilon_{(2i-1)^*}\}^\T \\\\
&= W_1+W_2+W_3+W_4.
\end{split}
\end{equation}
Thus, by the triangle inequality, we have 
\begin{equation}
\label{Eq:main concentration}
\|\hat{T}-T\|_{\max} = \|W_1\|_{\max}+\|W_2\|_{\max}+\|W_3\|_{\max}+\|W_4-T\|_{\max},
\end{equation}
where $\|W_1\|_{\max} = \max_{j,k} |(W_{1})_{jk}|$ is the largest absolute element in $W_1$.
It suffices to show that the $(j,k)$th element of the above terms are bounded above.

For sufficiently large $n$, we have 
\begin{equation}
\label{Eq:W1jk}
\begin{split}
(W_{1})_{jk} &=  \frac{1}{n} \sum_{i=1}^{\lfloor n/2 \rfloor} [m_j\{y_{(2i)}\}-m_j\{y_{(2i-1)}\}]  [m_k\{y_{(2i)}\}-m_k\{y_{(2i-1)}\}] \\
&\le n^{-1/2}  \left[ \frac{1}{n^{1/4}}  \underset{  \Xi_n (B)}{\sup}  \sum_{i=2}^{n}\|m\{y_{(i)}\}-m\{y_{(i-1)}\}\|_{\infty}\right]^2\\
&\le  \tau^2  n^{-1/2},
\end{split}
\end{equation}
where the last inequality holds by Assumption~\ref{ass:smoothness} for some arbitrary small constant  $\tau>0$.    Since this upper bound hold uniformly for all $j,k$, we have $\|W_1\|_{\max} \le \tau^2 n^{-1/2} $.

We now obtain an upper bound for $W_2$. We use the fact that  $E(\epsilon_{ij}) = E(x_{ij})- E\{E(x_{ij}\mid y_i)\} = 0$.  By Proposition~\ref{prop:property}, $\epsilon_{ij}$ is sub-Gaussian with mean zero and sub-Gaussian norm $2\|x_j\|_{\psi_2}$. Suppose that $2\|x_j\|_{\psi_2}\le L$ for some constant $L>0$.  
Thus, by Lemma~\ref{lemma:tail subgaussian}, there exists constant $C$ such that  $\mathrm{pr}\{\max_{i,j} \; |\epsilon_{ij}| \ge C(\log d)^{1/2}\} \le \exp ( -C' \log d  )$ with $C'>2$.
We have with probability at least $1- \exp ( -C' \log d  )$, 
\begin{equation}
\label{Eq:W2jk}
\begin{split}
|(W_{2})_{jk}| &= \left|\frac{1}{n} \sum_{i=1}^{\lfloor n/2 \rfloor} \{\epsilon_{(2i)^*,j}-\epsilon_{(2i-1)^*,j}\} [m_k\{y_{(2i)}\}-m_k\{y_{(2i-1)}\}]^\T\right| \\
&\le  \frac{2}{n^{3/4}}  \max_{i,j}\;| \epsilon_{ij}    |     \left[\frac{1}{n^{1/4}}  \underset{ \Xi_n (B)}{\sup}  \sum_{i=2}^{n}\|m\{y_{(i)}\}-m\{y_{(i-1)}\}\|_{\infty}\right]\\
&\le  C  \frac{1}{n^{1/4}}  (\log d/n)^{1/2} \tau,
\end{split}
\end{equation}
where the last inequality follows from Assumption~\ref{ass:smoothness} for some arbitrarily small constant $\tau>0$. 
By taking the union bound, we have 
\begin{equation*}
\begin{split}
\mathrm{pr} \left\{   \|W_2\|_{\max}  \ge C  \frac{1}{n^{1/4}}  (\log d/n)^{1/2} \tau \right\} &\le 
\sum_{j,k} \mathrm{pr} \left\{   |(W_{2})_{jk}|  \ge C  \frac{1}{n^{1/4}}  (\log d/n)^{1/2} \tau \right\}\\
&\le d^2 \mathrm{pr} \left\{   |(W_{2})_{jk}|  \ge C  \frac{1}{n^{1/4}}  (\log d/n)^{1/2} \tau \right\}\\
&\le  \exp (-C' \log d + 2\log d)\\
&= \exp (-C'' \log d).
\end{split}
\end{equation*}
Thus, we have $\|W_2\|_{\max} \le C  n^{-1/4} (\log d/n)^{1/2} \tau$ with probability at least $1- \exp ( -C'' \log d  )$.
The term $\|W_{3}\|_{\max}$ can be upper bounded similarly.

We now provide an upper bound on the term $\|W_{4} - T\|_{\max}$.  We have 
\begin{equation}
\label{Eq:W4jk}
\begin{split}
(W_{4})_{jk}-T_{jk} &=  \frac{1}{n} \sum_{i=1}^{\lfloor n/2 \rfloor}  \{\epsilon_{(2i)^*,j}-\epsilon_{(2i-1)^*,j}\} \{\epsilon_{(2i)^*,k}-\epsilon_{(2i-1)^*,k}\} - T_{jk}\\
&=\left( \frac{1}{n} \sum_{i=1}^n \epsilon_{ij}\epsilon_{ik} - T_{jk} \right)- \frac{1}{n} \sum_{i=1}^{\lfloor n/2 \rfloor} \epsilon_{(2i)^*,j} \epsilon_{(2i-1)^*,k}
-\frac{1}{n} \sum_{i=1}^{\lfloor n/2 \rfloor} \epsilon_{(2i)^*,k} \epsilon_{(2i-1)^*,j}.
\end{split}
\end{equation}
By Lemma~\ref{lemma:subgaussian}, $\epsilon_{ij}\epsilon_{ik} - T_{jk}$ is sub-exponential random variable with mean zero and with sub-exponential norm bounded by $C L^2$, since $\|x_j\|_{\psi_2} \le L$ for some constant $L> 0$.  By Lemma~\ref{lemma:bernstein}, we obtain
\[
\mathrm{pr} \left(\left|  \frac{1}{n} \sum_{i=1}^n \epsilon_{ij}\epsilon_{ik} - T_{jk} \right|\ge t \right) \le 2 \exp \left\{ -C'  \min \left( \frac{nt^2}{C^2 L^4}, \frac{nt}{CL^2}\right)\right\}.
\]
Similar to the proof of upper bound for $\|W_2\|_{\max}$, taking the union bound and picking $t = C  (\log d/n)^{1/2}$ for some sufficiently large constant $C$, we have 
\[
\max_{j,k}\left|  \frac{1}{n} \sum_{i=1}^n \epsilon_{ij}\epsilon_{ik} - T_{jk} \right|\le  C (\log d /n)^{1/2},
\]
with probability at least $1-\exp (-C' \log d )$.

It remains to show that the rest of the terms in (\ref{Eq:W4jk}) is upper bounded by $C  (\log d / n)^{1/2}$. 
Throughout the rest of the argument, we conditioned on the event that the order statistics are given.  Given the order statistics $y_{(1)},\ldots,y_{(n)}$, the concomitants $\epsilon_{(2i)^*,j}$ and  $\epsilon_{(2i-1)^*,k}$ are independent with mean zero \citep{yang1977general}.
Thus, by Lemma~\ref{lemma:subgaussian}, $\epsilon_{(2i)^*,j} \epsilon_{(2i-1)^*,k}$ is sub-exponential with mean zero and sub-exponential norm upper bounded by $CL^2$.  By Lemma~\ref{lemma:bernstein}, taking the union bound, and picking $t  = C (\log d /n)^{1/2}$, we have 
\[
\mathrm{pr} \left\{ \max_{j,k}\left| \frac{1}{n} \sum_{i=1}^{\lfloor n/2 \rfloor} \epsilon_{(2i)^*,j} \epsilon_{(2i-1)^*,k} \right|\ge C(\log d/n)^{1/2} \mid y_{(1)},\ldots,y_{(n)}\right\} \le  \exp ( - C' \log d)  .
\]
Since the above expression holds for any order statistics of $y$, we have 
\begin{equation*}
\begin{split}
&\mathrm{pr} \left\{ \max_{j,k}\left| \frac{1}{n} \sum_{i=1}^{\lfloor n/2 \rfloor} \epsilon_{(2i)^*,j} \epsilon_{(2i-1)^*,k} \right|\ge C(\log d/n)^{1/2} \right\}\\ 
&= E\left[\mathrm{pr} \left\{ \max_{j,k}\left| \frac{1}{n} \sum_{i=1}^{\lfloor n/2 \rfloor} \epsilon_{(2i)^*,j} \epsilon_{(2i-1)^*,k} \right|\ge C(\log d/n)^{1/2} \mid y_{(1)},\ldots,y_{(n)}\right\}\right]\\
&\le   \exp ( - C' \log d),
\end{split}
\end{equation*}
which implies
\[
 \max_{j,k}\left| \frac{1}{n} \sum_{i=1}^{\lfloor n/2 \rfloor} \epsilon_{(2i)^*,j} \epsilon_{(2i-1)^*,k} \right| \le C(\log d/n)^{1/2}
\] 
with probability at least $1-\exp ( - C' \log d)$. Thus, we have $\|W_4 - T\|_{\max} \le C(\log d/n)^{1/2}$ with probability at least $1-\exp ( - C' \log d)$.

Combining the upper bounds in (\ref{Eq:W1jk})-(\ref{Eq:W4jk}), we have 
\[
\|\hat{T}-T\|_{\max} \le n^{-1/2} \tau^2 + Cn^{-1/4}\tau (\log d /n)^{1/2}  + C  (\log d /n)^{1/2} \le C' (\log d /n)^{1/2}
\]
for sufficiently large $n$, since $(\log d /n)^{1/2}$ is the dominating term.

%%%%%%%%%%%%%%%%%%%%
% Proof of Estimation Error
%%%%%%%%%%%%%%%%%%%%
\section{Proof of Theorem~\ref{theorem:upper bound}}
\label{appendix:theorem 1}
The proof of Theorem~\ref{theorem:upper bound} is motivated from \citet{gao2014sparse} in the context of sparse canonical correlation analysis. The proofs of the technical lemmas in this section is deferred to Section~\ref{appendix:proof of lemma1}.

We begin with defining some notation that will be used throughout the proof of Theorem~\ref{theorem:upper bound}.
Let $\mathcal{S}_v$ be the set containing indices of non-zero rows of $V\in \RR^{d\times K}$.  Thus, $|\mathcal{S}_v| = s$.
Also, recall that $\Pi = V V^\T$ and let $\mathcal{S} =\mathrm{supp}(\Pi)$.  Thus, $|\mathcal{S}| \le s^2$.
Let $\mathcal{S}^c$ be the complementary set of $\mathcal{S}$.  
Let $\hat{\Pi}$ be a solution from solving (\ref{Eq:variational3}).  Let $\mathcal{V}$ and $\hat{\mathcal{V}}$ be the subspaces for $\Pi$ and $\hat{\Pi}$.  The goal is to establish the rate of convergence for the subspace distance $D(\mathcal{V}, \hat{\mathcal{V}})$.  
The following proposition reparametrizes the conditional covariance matrix $\Sigma_{E(x\mid y)}$ in terms of $V$, $\Lambda$, and $\Sigma_{x}$.

%%%%%%%%%%%%%%%%%%%%%%%
% Proposition
%%%%%%%%%%%%%%%%%%%%%%%
\begin{proposition}
\label{prop:sol}
The solution, up to sign jointly, of (\ref{eq:1}) is $V$ if and only if $\Sigma_{E(x \mid y)}$ can be written as 
\[
\Sigma_{E(x \mid y)} = \Sigma_{x} V \Lambda V^\T \Sigma_{x},
\]
where $V^\T \Sigma_{x} V = I_K$. 
\end{proposition}

Note that $\hat{V}$ is normalized with respect to the sample covariance matrix $\hat{\Sigma}_{x}$.  In contrast, the truth $V $ is normalized with respect to $\Sigma_{x}$.
Motivated by \cite{gao2014sparse}, to facilitate the proof, we let 
\[
\tilde{\Sigma}_{E(x \mid y)} =  \hat{\Sigma}_{x} V \Lambda V^\T \hat{\Sigma}_{x}, \quad      \tilde{V} = V ( V^\T \hat{\Sigma}_{x} V)^{-1/2}, \quad \tilde{\Pi} = \tilde{V} \tilde{V}^\T,
\quad \mathrm{and} \quad  \tilde{\Lambda} = ( V^\T \hat{\Sigma}_{x} V)^{1/2} \Lambda ( V^\T \hat{\Sigma}_{x} V )^{1/2}. 
\]

The intuition of $ \tilde{V}$ and $\tilde{\Lambda}$ is to approximate $V$ and $\Lambda$, respectively, since $V^\T \hat{\Sigma}_{x} V$ is close to the  identity matrix $I_K$. 
The following lemma establishes concentration between $\tilde{\Lambda}$ and $\Lambda$, and between $\Pi$ and $\tilde{\Pi}$.

%%%%%%%%%%%%%%%%%%%%%%%%%%%%%%
% Lemma 
%%%%%%%%%%%%%%%%%%%%%%%%%%%%%%
\begin{lemma}
\label{lemma:difference}
For any $C>0$, there exists $C'>0$ depending only  on $C$ such that 
\[
\| \tilde{\Lambda} - \Lambda    \|_{\mathrm{sp}} \le C(s/n)^{1/2}
\quad \mathrm{and} 
\quad
\|\tilde{\Pi}-\Pi\|_\mathrm{F}\le C K (s/n)^{1/2},
\]
with probability greater than $1-\exp ( - C' s   )$.
\end{lemma}

Throughout the proof of Theorem~\ref{theorem:upper bound}, we let $\Delta = \hat{\Pi}-\tilde{\Pi}$. We further partition the set $\mathcal{S}^c$ into $J$ sets such that $\mathcal{S}_1^c$ is the index set of the largest $l$ entries of $|\Delta|$, $\mathcal{S}_2^c$ is the index set of the second largest $l$ entries of $|\Delta|$, and so forth, with $|\mathcal{S}_{J}^c| \le l$.
We first state some technical lemmas that are needed to prove Theorem~\ref{theorem:upper bound}.  
For two matrices $A,B$, we define the inner product as $\langle A,B\rangle = \mathrm{tr}(AB)$
 The following curvature lemma establishes a bound on the curvature of the objective function.   It follows directly from \citet{gao2014sparse}.

%%%%%%%%%%%%%%%%%%%%%%%%%%%%%%
% Lemma 
%%%%%%%%%%%%%%%%%%%%%%%%%%%%%%
\begin{lemma}
\label{lemma:curvature}
Let $U \in \RR^{d\times K}$ be an orthonormal matrix, i.e., $U^\T U = I_K$.  Furthermore, let $L\in \RR^{K\times K}$ and $D= \mathrm{diag}(d_1,\ldots,d_K) \in \RR^{K \times K}$, where $d_1 \ge \cdots \ge d_K > 0$. If $E \in \RR^{d\times d}$ is such that  $\|E\|_* \leq K$ and $\| E \|_\mathrm{sp} \leq 1$, then
\[
\bigl\langle U L U^\T, UU^\T - E \bigr\rangle \geq \frac{d_K}{2} \left\| UU^\T - E \right\|_{\mathrm{F}}^2 - \|L - D\|_\mathrm{F}  \|U U^\T - E\|_\mathrm{F}. 
\]
\end{lemma}

\noindent The following lemma shows that $\tilde{\Pi}$ satisfies the constrains 
\[
\bigl\| \hat{\Sigma}^{1/2}_{x} \tilde{\Pi} \hat{\Sigma}^{1/2}_{x} \bigr\|_* \leq K \qquad  \mathrm{and} \qquad \bigl\| \hat{\Sigma}^{1/2}_{x} \tilde{\Pi} \hat{\Sigma}^{1/2}_{x} \bigr\|_\mathrm{sp} \leq 1.
\]
%%%%%%%%%%%%%%%%%%%%%%%%%%%%%%
% Lemma
%%%%%%%%%%%%%%%%%%%%%%%%%%%%%%
\begin{lemma}
\label{lemma:optim}
Let $\tilde{V} = V ( V^\T \hat{\Sigma}_{x} V)^{-1/2}$ and let $\tilde{\Pi} = \tilde{V} \tilde{V}^\T$. 
Then, $\tilde{\Pi}$ satisfies both the  constrains  
\[
\bigl\| \hat{\Sigma}^{1/2}_{x} \tilde{\Pi} \hat{\Sigma}^{1/2}_{x} \bigr\|_* \leq K \qquad  \mathrm{and} \qquad \bigl\| \hat{\Sigma}^{1/2}_{x} \tilde{\Pi} \hat{\Sigma}^{1/2}_{x} \bigr\|_\mathrm{sp} \leq 1.
\]
\end{lemma}

%%%%%%%%%%%%%%%%%%%%%%%%%%%%%%
% Lemma
%%%%%%%%%%%%%%%%%%%%%%%%%%%%%%
\begin{lemma}
\label{lemma:complement set}
We have 
\[
\sum_{j=2}^J \|\Delta_{\mathcal{S}_j^c}\|_\mathrm{F} \le l^{-1/2} \| \Delta_{\mathcal{S}^c}  \|_1.
\]
\end{lemma}

%%%%%%%%%%%%%%%%%%%%%%%%%%%%%%
% Lemma
%%%%%%%%%%%%%%%%%%%%%%%%%%%%%%
\begin{lemma}
\label{lemma:sample eigenvalue}
 Assume that $x_1,\ldots,x_n$ are independent sub-Gaussian random variables  with covariance matrix $\Sigma_x$.  
Let $\hat{\Sigma}_{x}$ be the sample covariance matrix.  Let $s < \min (n,d)$.  For any $C>0$, there exists a constant $C'>0$ such that 
\[
c^{-1} - C \{s\log (ed)/n\}^{1/2} \le \lambda_{\min} (\hat{\Sigma}_{x},s)\le  \lambda_{\max} (\hat{\Sigma}_{x},s)\le  c + C \{s\log (ed)/n\}^{1/2},
\]
with probability greater than $1-\exp\{-C' s\log(ed)\}$, where $c$ is a constant from Assumption~\ref{ass:bounded eigenvalues}.
\end{lemma}

Finally, we need a lemma on the concentration between $\tilde{\Sigma}_{E(x \mid y)}$ and $\hat{\Sigma}_{E(x \mid y)}$.
\begin{lemma}
\label{lemma:tilde conditional concentration}
Let $\hat{\Sigma}_{E(x \mid y)}$ be the estimator as defined in Corollary~\ref{cor:concentration} and let $\tilde{\Sigma}_{E(x \mid y)} =  \hat{\Sigma}_{x} V \Lambda V^\T \hat{\Sigma}_{x}$.  Assume that $K\lambda_1(\log d/n)^{1/2}$ is bounded by some constant. We have 
\[
\|\tilde{\Sigma}_{E(x \mid y)}- \hat{\Sigma}_{E(x \mid y)}\|_{\max} \le  C (\log d/n)^{1/2}
\]
with probability at least $1-\exp (-C' \log  d )$.
\end{lemma}

We now provide the proof of Theorem~\ref{theorem:upper bound}. 

%%%%%%%%%%%%%%%%%%%%%%%%%%%%%
%%%%%%%%%%%%%%%%%%%%%%%%%%%%%%
% Proof of Theorem upper bound
%%%%%%%%%%%%%%%%%%%%%%%%%%%%%
%%%%%%%%%%%%%%%%%%%%%%%%%%%%%%
\begin{proof}
Recall that $\Delta = \hat{\Pi}-\tilde{\Pi}$.
The proof involves obtaining an upper bound and a lower bound on the quantity $\|\hat{\Sigma}^{1/2}_{x}\Delta\hat{\Sigma}^{1/2}_{x}\|_\mathrm{F}^2$.  Combining the upper and lower bounds, we obtain an upper bound on $\|\Delta\|_\mathrm{F}$.

Upper bound for $\|\hat{\Sigma}^{1/2}_{x}\Delta\hat{\Sigma}^{1/2}_{x}\|_\mathrm{F}^2$:
By Lemma~\ref{lemma:optim}, $\tilde{\Pi}$ is a feasible solution of  (\ref{Eq:variational3}).  Since $\hat{\Pi}$ is the optimum solution of (\ref{Eq:variational3}), we have $- \langle \hat{\Sigma}_{E(x \mid y)} , \hat{\Pi} \rangle + \rho \| \hat{\Pi} \|_{1} \leq - \langle \hat{\Sigma}_{E(x \mid y)}, \tilde{\Pi} \rangle + \rho\| \tilde{\Pi} \|_{1}$,
implying 
\begin{equation}
\label{Eq:theoremupper1}
\begin{split}
- \langle \tilde{\Sigma}_{E(x \mid y)} , \Delta \rangle  &\le \langle \hat{\Sigma}_{E(x \mid y)}-\tilde{\Sigma}_{E(x \mid y)}, \Delta \rangle + \rho\| \tilde{\Pi} \|_{1}-\rho \| \hat{\Pi} \|_{1}\\
&\le \|\hat{\Sigma}_{E(x \mid y)}-\tilde{\Sigma}_{E(x \mid y)}\|_{\max} \|\Delta\|_{1} + \rho\| \tilde{\Pi} \|_{1}-\rho \| \tilde{\Pi} + \Delta \|_{1}\\
&\le \frac{\rho}{2} \|\Delta\|_{1} + \rho\| \tilde{\Pi} \|_{1}-\rho \| \tilde{\Pi} + \Delta \|_{1},
\end{split}
\end{equation}
where the second inequality holds by Holder's inequality, and the last inequality holds by picking $\rho> 2 \|\tilde{\Sigma}_{E(x \mid y)} - \hat{\Sigma}_{E(x \mid y)}\|_{\max}$.  
By definition of $\tilde{\Pi}$, we have $\mathrm{supp}(\tilde{\Pi})= \mathrm{supp}(\Pi)$. Thus, we obtain
\begin{equation}
\label{Eq:theoremupper2}
\|\tilde{\Pi}\|_{1} - \|\tilde{\Pi} + \Delta \|_{1}
= \|\tilde{\Pi}_{\mathcal{S}}\|_{1} - \|\tilde{\Pi}_\mathcal{S}+ \Delta_{\mathcal{S}} \|_{1}- \|\Delta_{\mathcal{S}^c} \|_{1}\le  \|\Delta_{\mathcal{S}} \|_{1} - \|\Delta_{\mathcal{S}^c} \|_{1},
\end{equation}
where the last inequality follows from the triangle inequality.  Substituting (\ref{Eq:theoremupper2}) into (\ref{Eq:theoremupper1}), we obtain
\begin{equation}
\label{Eq:theoremupper3}
- \langle \tilde{\Sigma}_{E(x \mid y)} , \Delta \rangle \leq \frac{3\rho}{2} \| \Delta_{\mathcal{S}} \|_{1} - \frac{\rho}{2} \| \Delta_{\mathcal{S}^c}\|_{1}.
\end{equation}

%%%%%%%%%%%%%%%%%%%%%%%%%%%         
%%% Lower bound 
%%%%%%%%%%%%%%%%%%%%%%%%%%% 
Next, we obtain a lower bound for $- \langle \tilde{\Sigma}_{E(x \mid y)} , \Delta \rangle$.
By an application of Lemma~\ref{lemma:curvature}, we have 
\begin{equation}
\label{Eq:theoremupper4}
\begin{split}
- \langle \tilde{\Sigma}_{E(x \mid y)} , \Delta \rangle &= \langle \hat{\Sigma}^{1/2}_{x} V \Lambda V^\T \hat{\Sigma}^{1/2}_{x}, \hat{\Sigma}^{1/2}_{x} (\tilde{\Pi} - \hat{\Pi}) \hat{\Sigma}^{1/2}_{x} \rangle\\
& = \langle \hat{\Sigma}^{1/2}_{x} \tilde{V} \tilde{\Lambda} \tilde{V}^{\T} \hat{\Sigma}^{1/2}_{x}, \hat{\Sigma}^{1/2}_{x} (\tilde{\Pi} - \hat{\Pi}) \hat{\Sigma}^{1/2}_{x} \rangle \\
&\geq \frac{\lambda_K}{2} \| \hat{\Sigma}^{1/2}_{x} (\tilde{\Pi} - \hat{\Pi}) \hat{\Sigma}^{1/2}_{x} \|_\mathrm{F}^2 -\|\tilde{\Lambda}-\Lambda\|_\mathrm{F}\| \hat{\Sigma}^{1/2}_{x} (\tilde{\Pi} - \hat{\Pi}) \hat{\Sigma}^{1/2}_{x} \|_\mathrm{F},
\end{split}
\end{equation}
where $\lambda_K$ is the $K$th generalized eigenvalue of the pair of matrices $\{\Sigma_{E(x \mid y)}, \Sigma_{x}\}$.
For notational convenience, we let $\gamma = \|\tilde{\Lambda}-\Lambda\|_\mathrm{F}$.
Combining (\ref{Eq:theoremupper4}) and (\ref{Eq:theoremupper3}), we obtain
\begin{equation}
\label{Eq:theoremupper4-5}
\lambda_K\| \hat{\Sigma}^{1/2}_{x} \Delta \hat{\Sigma}^{1/2}_{x} \|_\mathrm{F}^2 
\le 2\gamma   \| \hat{\Sigma}^{1/2}_{x} \Delta \hat{\Sigma}^{1/2}_{x} \|_\mathrm{F} 
+ 3\rho\| \Delta_{\mathcal{S}} \|_{1} - \rho \| \Delta_{\mathcal{S}^c}\|_{1},
\end{equation}
implying 
\begin{equation}
\label{Eq:theoremupper5}
\| \hat{\Sigma}^{1/2}_{x} \Delta \hat{\Sigma}^{1/2}_{x} \|_\mathrm{F}^2 
\le \frac{2\gamma}{\lambda_K}   \| \hat{\Sigma}^{1/2}_{x} \Delta \hat{\Sigma}^{1/2}_{x} \|_\mathrm{F} 
+ \frac{3\rho}{\lambda_K}\| \Delta_{\mathcal{S}} \|_{1}.
\end{equation}
Thus, 
\begin{equation}
\label{Eq:theoremupper6}
\| \hat{\Sigma}^{1/2}_{x} \Delta \hat{\Sigma}^{1/2}_{x} \|_\mathrm{F}^2 
\le \frac{4\gamma^2}{\lambda_K^2}
+ \frac{6\rho}{\lambda_K}\| \Delta_{\mathcal{S}} \|_{1},
\end{equation}
where we use the fact that  $ax^2\le bx+c$ implies that $x^2 \le {b^2}/{a^2}+ {2c}/{a}$.

%%%%%%%%%%%%%%%%%%%%%%%%%%
% Lower bound
%%%%%%%%%%%%%%%%%%%%%%%%%%
Lower bound for $\|\hat{\Sigma}^{1/2}_{x}\Delta\hat{\Sigma}^{1/2}_{x}\|_\mathrm{F}^2$:  
We start by showing that $\Delta$  lies in a restricted set, referred to as the generalized cone condition in \citet{gao2014sparse}.
By (\ref{Eq:theoremupper4-5}), we have 
\begin{equation*}
\begin{split}
0
&\le 2\gamma   \| \hat{\Sigma}^{1/2}_{x} \Delta \hat{\Sigma}^{1/2}_{x} \|_\mathrm{F} 
+ 3\rho\| \Delta_{\mathcal{S}} \|_{1} - \rho \| \Delta_{\mathcal{S}^c}\|_{1}\\
&\le \frac{\gamma^2}{\lambda_K}  + \lambda_K \| \hat{\Sigma}^{1/2}_{x} \Delta \hat{\Sigma}^{1/2}_{x} \|_\mathrm{F}^2 
+ 3\rho\| \Delta_{\mathcal{S}} \|_{1} - \rho \| \Delta_{\mathcal{S}^c}\|_{1}\\
&\le \frac{5\gamma^2}{\lambda_K} 
+ 9\rho\| \Delta_{\mathcal{S}} \|_{1} - \rho \| \Delta_{\mathcal{S}^c}\|_{1},
\end{split}
\end{equation*}
where the second inequality is obtained by using the fact that $2ab \le a^2 + b^2$, and the last inequality is obtained by substituting (\ref{Eq:theoremupper6}) into the second expression.
This implies that 
\begin{equation}
\label{Eq:theoremupper7}
 \| \Delta_{\mathcal{S}^c}\|_{1} \le 9\| \Delta_{\mathcal{S}} \|_{1} +  \frac{5\gamma^2}{\rho \lambda_K}.
\end{equation}
Furthermore, by Lemma~\ref{lemma:complement set} and (\ref{Eq:theoremupper7}), we have
\begin{equation}
\label{Eq:theoremupper8}
\sum_{j=2}^J \|\Delta_{\mathcal{S}_j^c}\|_\mathrm{F} \le l^{-1/2} \| \Delta_{\mathcal{S}^c}  \|_1
\le 9l^{-1/2} \|\Delta_{\mathcal{S}}\|_{1} + \frac{5\gamma^2}{ \rho \lambda_K  l^{1/2}}\le  9sl^{-1/2} \|\Delta_{\mathcal{S}}\|_{\mathrm{F}} + \frac{5\gamma^2}{ \rho \lambda_K l^{1/2} },
\end{equation}
where the last inequality is obtained by using the fact that $\|\Delta_\mathcal{S}\|_1 \le s \|\Delta_{\mathcal{S}}\|_\mathrm{F}$.

Thus, by the triangle inequality, we have 
\begin{equation}
\label{Eq:theoremupper9}
\begin{split}
\|\hat{\Sigma}^{1/2}_{x} \Delta \hat{\Sigma}^{1/2}_{x} \|_\mathrm{F} &\geq \|\hat{\Sigma}^{1/2}_{x} \Delta_{\mathcal{S}\cup \mathcal{S}^c_1} \hat{\Sigma}^{1/2}_{x} \|_\mathrm{F} - \sum_{j=2}^J \|\hat{\Sigma}^{1/2}_{x} \Delta_{\mathcal{S}_j^c} \hat{\Sigma}^{1/2}_{x} \|_\mathrm{F} \\
&\geq  \lambda_{\min} (\hat{\Sigma}_{x},s+l)    \| \Delta_{\mathcal{S}\cup \mathcal{S}^c_1} \|_\mathrm{F} -  \lambda_{\max} (\hat{\Sigma}_{x},l)    \sum_{j=2}^J \| \Delta_{ \mathcal{S}^c_j} \|_\mathrm{F},
\end{split}
\end{equation}
where $ \lambda_{\min} (\hat{\Sigma}_{x},s+l)$ and $\lambda_{\max} (\hat{\Sigma}_{x},l) $ are the minimum $(s+l)$-sparse eigenvalue and maximum  $l$-sparse eigenvalue of $\hat{\Sigma}_{x}$, respectively. 
Substituting (\ref{Eq:theoremupper8}) into (\ref{Eq:theoremupper9}), we have 
\begin{equation}
\label{Eq:theoremupper10}
\|\hat{\Sigma}^{1/2}_{x} \Delta \hat{\Sigma}^{1/2}_{x} \|_\mathrm{F} \ge \left\{ \lambda_{\min} (\hat{\Sigma}_{x},s+l) -9\lambda_{\max} (\hat{\Sigma}_{x},l) s l^{-1/2} \right\} \| \Delta_{\mathcal{S}\cup \mathcal{S}^c_1} \|_\mathrm{F} - \frac{5\lambda_{\max} (\hat{\Sigma}_{x},l)\gamma^2}{\rho \lambda_K l^{1/2}}.
\end{equation}

Choose $l=c_1s^2$.  By Lemma S5, we have with probability at least $1-\exp\{C' (c_1 s^2+s)\log {ed}\}$,
\[
c^{-1} - C \left\{\frac{(c_1s^2 +s)\log (ed)}{n} \right\}^{1/2} \le  \lambda_{\min} (\hat{\Sigma}_{x},s+l)\le   \lambda_{\max} (\hat{\Sigma}_{x},s+l) \le c+ C \left\{\frac{(c_1s^2+s) \log (ed)}{n} \right\}^{1/2}. 
\]
Thus, we have 
\begin{equation*}
\begin{split}
& \lambda_{\min} (\hat{\Sigma}_{x},s+l) -9\lambda_{\max} (\hat{\Sigma}_{x},l) s l^{-1/2} \\
 &=  \lambda_{\min} (\hat{\Sigma}_{x},s+l) -9c_1^{-1/2}\lambda_{\max} (\hat{\Sigma}_{x},l) \\
 &\ge c^{-1}- C\left\{{\frac{(c_1s^2+s)\log (ed)}{n}}\right\}^{1/2}  - 9c_1^{-1/2} c - 9c_1^{-1/2}C\left\{\frac{(c_1s^2+s)\log (ed)}{n}\right\}^{1/2}.
 \end{split}
\end{equation*}
This quantity can be lower bounded by a constant $C_1>0$ as long as $c_1$ is sufficiently large and that $n >  C_2 (c_1s^2+s)\log (ed)$. 
Similarly, the term $5\lambda_{\max} (\hat{\Sigma}_{x},l)$ can be upper bounded by a constant $C_3>0$ by Assumption~\ref{ass:bounded eigenvalues}. Thus, we have
\begin{equation}
\label{Eq:theoremupper11}
\|\hat{\Sigma}^{1/2}_{x} \Delta \hat{\Sigma}^{1/2}_{x} \|_\mathrm{F}  \ge C_1 \| \Delta_{\mathcal{S}\cup \mathcal{S}^c_1} \|_\mathrm{F} - C_3 \frac{ \gamma^2}{ \rho \lambda_K l^{1/2}}.
\end{equation}

%%%%%%%%%%%%%%%%%%%%%%%%%%
% Combining
%%%%%%%%%%%%%%%%%%%%%%%%%%
Combining the lower and upper bounds for $\|\hat{\Sigma}^{1/2}_{x} \Delta \hat{\Sigma}^{1/2}_{x} \|_\mathrm{F}$: 
By (\ref{Eq:theoremupper11}) and (\ref{Eq:theoremupper6}), we obtain
\begin{equation*}
\begin{split}
C_1 \|\Delta_{\mathcal{S}\cup \mathcal{S}^c_1}\|_\mathrm{F} &\le C_3 \frac{ \gamma^2}{\rho \lambda_Kl^{1/2}}  +\|\hat{\Sigma}^{1/2}_{x} \Delta \hat{\Sigma}^{1/2}_{x} \|_\mathrm{F}   \\
&\le C_3\frac{ \gamma^2}{\rho \lambda_Kl^{1/2}}  + \left(\frac{4\gamma^2}{\lambda_K^2}
+ \frac{6\rho}{\lambda_K}\| \Delta_{\mathcal{S}} \|_{1}\right)^{1/2}\\
&\le C_3 \frac{ \gamma^2}{\rho \lambda_Kl^{1/2}}  +\left(\frac{4\gamma^2}{\lambda_K^2}
+ \frac{6\rho s}{\lambda_K}\| \Delta_{\mathcal{S}\cup \mathcal{S}_1^c} \|_{\mathrm{F}}\right)^{1/2},
\end{split}
\end{equation*}
where we use the fact that  $\|\Delta_{\mathcal{S}}\|_1 \le s\|\Delta_{\mathcal{S}}\|_\mathrm{F} \le s \|\Delta_{\mathcal{S} \cup \mathcal{S}_1^c }\|_\mathrm{F} $.
This implies that
\[
 \|\Delta_{\mathcal{S}\cup \mathcal{S}^c_1}\|_\mathrm{F}  \le C_4 \frac{ \gamma^2}{\rho \lambda_Kl^{1/2}}  +C_5\left(\frac{4\gamma^2}{\lambda_K^2}+  \frac{6\rho s}{\lambda_K}\| \Delta_{\mathcal{S}\cup \mathcal{S}_1^c} \|_{\mathrm{F}}\right)^{1/2}.
\]

 By squaring both sides, we have
 \begin{equation}
 \small
\label{Eq:theoremupper12}
 \begin{split}
 \|\Delta_{\mathcal{S}\cup \mathcal{S}^c_1}\|_\mathrm{F}^2 &\le C_4^2\frac{ \gamma^4}{l \rho^2 \lambda_K^2}  +C_5^2 \left(\frac{4\gamma^2}{\lambda_K^2}+  \frac{6\rho s}{\lambda_K}\| \Delta_{\mathcal{S}\cup \mathcal{S}_1^c} \|_{\mathrm{F}}\right) + 2 C_4C_5  \frac{ \gamma^2}{\rho \lambda_Kl^{1/2}} \left(\frac{4\gamma^2}{\lambda_K^2}+  \frac{6\rho s}{\lambda_K}\| \Delta_{\mathcal{S}\cup \mathcal{S}_1^c} \|_{\mathrm{F}}\right)^{1/2}\\
 &\le 2C_4^2\frac{ \gamma^4}{l \rho^2 \lambda_K^2}  +2C_5^2 \left(\frac{4\gamma^2}{\lambda_K^2}+  \frac{6\rho s}{\lambda_K}\| \Delta_{\mathcal{S}\cup \mathcal{S}_1^c} \|_{\mathrm{F}}\right)\\
 &=  2C_4^2\frac{ \gamma^4}{l \rho^2 \lambda_K^2}  +8C_5^2 \frac{\gamma^2}{\lambda_K^2}+ 12C_5^2   \frac{\rho s}{\lambda_K}\| \Delta_{\mathcal{S}\cup \mathcal{S}_1^c} \|_{\mathrm{F}},
 \end{split}
\end{equation}
where the second inequality holds by using the fact that $2ab \le a^2 + b^2$.
Using the fact that $ax^2\le bx+c$ implies that $x^2 \le {b^2}/{a^2}+ {2c}/{a}$, we obtain
\begin{equation}
\label{Eq:theoremupper13}
 \|\Delta_{\mathcal{S}\cup \mathcal{S}^c_1}\|_\mathrm{F}^2 \le C_6 \left( \frac{\gamma^2}{\lambda_K^2}+ \frac{ \gamma^4}{l \rho^2 \lambda_K^2}  +  \frac{\rho^2 s^2}{\lambda_K^2}\right).
\end{equation}

By the triangle inequality, $\|\Delta \|_\mathrm{F} \le  \|\Delta_{\mathcal{S}\cup \mathcal{S}^c_1}\|_\mathrm{F} + \|\Delta_{(\mathcal{S}\cup \mathcal{S}^c_1)^c}\|_\mathrm{F} \le \|\Delta_{\mathcal{S}\cup \mathcal{S}^c_1}\|_\mathrm{F}+\sum_{j=2}^J \|\Delta_{\mathcal{S}_j^c}\|_\mathrm{F} $.
Thus, by (\ref{Eq:theoremupper8}), we have
\begin{equation}\label{Eq:theoremupper14}
\|\Delta\|_\mathrm{F} \le \|\Delta_{\mathcal{S}\cup \mathcal{S}^c_1}\|_\mathrm{F} + 9sl^{-1/2} \|\Delta_{\mathcal{S}\cup \mathcal{S}^c_1}\|_\mathrm{F} + \frac{5\gamma^2}{\rho \lambda_K l^{1/2}}.
\end{equation}
Recall that $l = c_1 s^2$ and that $K<\log d$.  By Lemma~\ref{lemma:difference}, $\gamma = \|\tilde{\Lambda}-\Lambda\|_\mathrm{F}\le K^{1/2}\|\tilde{\Lambda}-\Lambda\|_\mathrm{sp} \le C (Ks/n)^{1/2} \le C_6 \rho l^{1/2}$, with probability at least $1-\exp(-C' s)$.
Substituting (\ref{Eq:theoremupper13}) into (\ref{Eq:theoremupper14}), we obtain
\begin{equation}
\label{Eq:theoremupper15}
\begin{split}
\|\Delta\|_\mathrm{F} &\le \left(1+\frac{9}{c_1} \right) \|\Delta_{\mathcal{S}\cup \mathcal{S}^c_1}\|_\mathrm{F} + \frac{5C_6 \rho l^{1/2}}{\lambda_K}\\
&\le  \left(1+\frac{9}{c_1} \right) \left\{C_6 \left( \frac{\gamma^2}{\lambda_K^2}+ \frac{ \gamma^4}{l \rho^2 \lambda_K^2}  +  \frac{\rho^2 s^2}{\lambda_K^2}\right)\right\}^{1/2} + \frac{5C_6 \rho l^{1/2}}{\lambda_K}\\
&=  \left(1+\frac{9}{c_1} \right) \left\{C_6 \left( \frac{C_6^2 \rho^2 c_1 s^2}{\lambda_K^2}+ \frac{ C_6^4 \rho^2 c_1s^2}{ \lambda_K^2}  +  \frac{\rho^2 s^2}{\lambda_K^2}\right)\right\}^{1/2} + \frac{5C_6 \rho c_1^{1/2} s }{\lambda_K}\\
&\le C \frac{s\rho}{\lambda_K},
\end{split}
\end{equation}
for $C$ sufficiently large, with large probability.

By the triangle inequality, we have 
\[
\|\hat{\Pi}-\Pi\|_\mathrm{F} \le \|\Delta\|_\mathrm{F} + \|\Pi-\tilde{\Pi}\|_\mathrm{F} \le  C\frac{s\rho}{\lambda_K} +  C K (s/n)^{1/2}%\le C \frac{s\rho}{\lambda_K},
\]
where the second inequality holds by an application of Lemma~\ref{lemma:difference}.

Let $\hat{\mathcal{V}}$ and $\mathcal{V}$ be the subspace corresponding to $\hat{\Pi}$ and $\Pi$, respectively.  
Then, by Corollary 3.2 of \citet{vu2013fantope}, we have 
\[
D\bigl(\mathcal{V}, \hat{\mathcal{V}}\bigr) \leq C \frac{s\rho}{\lambda_K}+ C K (s/n)^{1/2},
\]
for some constant $C$, which concludes the proof. Finally, by an application of Lemma~\ref{lemma:tilde conditional concentration}, we have $\rho \ge C(\log d/n)^{1/2}$  with probability at least $1- \exp (C' \log p )$.  Therefore 
\[
D\bigl(\mathcal{V}, \hat{\mathcal{V}}\bigr) \leq C \frac{s}{\lambda_K}\left( \frac{\log d}{n}\right)^{1/2}+ C K (s/n)^{1/2} \le C \frac{s}{\lambda_K} \left(\frac{\log d}{n}\right)^{1/2},
\]  
where the second inequality holds by the assumption that $\lambda_KK^2/\{s\log(d)\}$ is upper bounded by a constant.  
This concludes the proof.

\end{proof}

%%%%%%%%%%%%%%%%%%%%%%%%%%%%%%%%%%%%
%%%%%%%%%%%%%%%%%%%%%%%%%%%%%%%%%%%%
% Proof of Technical Lemma 
%%%%%%%%%%%%%%%%%%%%%%%%%%%%%%%%%%%%
%%%%%%%%%%%%%%%%%%%%%%%%%%%%%%%%%%%%
\section{Proof of Lemmas in Appendix~\ref{appendix:theorem 1}}
\label{appendix:proof of lemma1}

%%%%%%%%%%%%%%%%%%%%%%%%
% Proof of Lemma difference
%%%%%%%%%%%%%%%%%%%%%%%%
\subsection{Proof of Lemma~\ref{lemma:difference}}
The proof of Lemma~\ref{lemma:difference} requires the following probabilistic bound on the operator norm and a lemma on perturbation bound of square root matrices. The following proposition is a special case of Remark 5.40 in an unpublished 2015 technical report, Vershynin, R. (arXiv:1011.3027).
\begin{proposition}%[Remark 5.40 in an unpublished 2015 technical report, Vershynin, R. (arXiv:1011.3027)]
\label{prop:zmm}
Let $Y$ be an $n\times s$ matrix whose rows are independent sub-Gaussian isotropic random vectors in $\mathbb{R}^s$. Let $\delta = C (s/n)^{1/2} + t/(n)^{1/2}$.   Then, for every $t\ge 0$,
\[
\mathrm{pr}\left\{\left\| \frac{1}{n}Y^\T Y - I_s  \right\|_\mathrm{sp}   \le 
\max (\delta,\delta^2)
\right\} \ge 1-2\exp(-ct^2/2).
\] 
The constants $c,C$ depend only on the sub-Gaussian norm of the rows.
\end{proposition}

\begin{lemma}[Lemma 2 in \citealp{gao2014minimax}]
\label{lemma:perturb}
Let $E$ and $F$ be positive semi-definite matrices.  Then, for any unitarily invariant norm $\|\cdot \|_\mathrm{sp}$, we have 
\[
\|E^{1/2}-F^{1/2}\|_\mathrm{sp} \le \frac{1}{\sigma_{\min}(E^{1/2})+\sigma_{\min}(F^{1/2})} \|E-F\|_\mathrm{sp},
\]	
where $\sigma_{\min} (E^{1/2})$ is the smallest non-zero singular value of $E^{1/2}$.
\end{lemma}

We are now ready to prove Lemma~\ref{lemma:difference}.
\begin{proof}
Recall from Appendix~\ref{appendix:theorem 1} that $\tilde{\Lambda} = (V^\T \hat{\Sigma}_{x} V)^{1/2} \Lambda (V^\T \hat{\Sigma}_{x} V )^{1/2}$.  Then
\begin{equation}
\label{Eq:lemmadifference1}
\begin{split}
\|\tilde{\Lambda}-\Lambda\|_\mathrm{sp} &=\left\| \bigl( V^\T \hat{\Sigma}_{x} V\bigr)^{1/2} \Lambda \bigl( V^\T \hat{\Sigma}_{x} V \bigr)^{1/2} - \Lambda \right\|_\mathrm{sp}\\
&= \left\| \bigl( V^\T \hat{\Sigma}_{x} V\bigr)^{1/2} \Lambda \bigl( V^\T \hat{\Sigma}_{x} V \bigr)^{1/2}   -\Lambda \bigl( V^\T \hat{\Sigma}_{x} V \bigr)^{1/2} +  \Lambda \bigl(V^\T \hat{\Sigma}_{x} V \bigr)^{1/2} -    \Lambda \right\|_\mathrm{sp}\\
&\le \left\| \bigl( V^\T \hat{\Sigma}_{x} V\bigr)^{1/2} \Lambda \bigl( V^\T \hat{\Sigma}_{x} V \bigr)^{1/2}   -\Lambda \bigl( V^\T \hat{\Sigma}_{x} V \bigr)^{1/2}\right\|_\mathrm{sp} + \left\|  \Lambda \bigl( V^\T \hat{\Sigma}_{x} V \bigr)^{1/2} -    \Lambda \right\|_\mathrm{sp}\\
&\le \|  \Lambda  \|_\mathrm{sp}  \left\| \bigl( V^\T \hat{\Sigma}_{x} V \bigr)^{1/2}-I_K    \right\|_\mathrm{sp} \left\| \bigl( V^\T \hat{\Sigma}_{x} V \bigr)^{1/2}    \right\|_\mathrm{sp} + \|\Lambda\|_\mathrm{sp} \left\| \bigl( V^\T \hat{\Sigma}_{x} V \bigr)^{1/2}-I_K    \right\|_\mathrm{sp}.
\end{split}
\end{equation}
By Lemma~\ref{lemma:perturb},  
\begin{equation}
\label{Eq:lemmadifference1-5}
  \left\| \bigl( V^\T \hat{\Sigma}_{x} V \bigr)^{1/2}-I_K    \right\|_\mathrm{sp}\le C \left\| V^\T \hat{\Sigma}_{x} V -I_K    \right\|_\mathrm{sp}
 \end{equation}
 for some constant $C$.
 Thus, it suffices to establish an upper bound on $ \|  V^\T \hat{\Sigma}_{x} V -I_K    \|_\mathrm{sp}$.
Recall that $\hat{\Sigma}_{x} = {X}^\T{X}/n$, where each row of $X$ is independent sub-Gaussian random variables with covariance matrix $\Sigma_x$.  Also, recall that $|\mathcal{S}_v| =  s$.
Thus, by the definition of the spectral norm, we obtain
\begin{equation}
\label{Eq:lemmadifference2}
\begin{split}
 \left\| V^\T \hat{\Sigma}_{x} V -I_K    \right\|_\mathrm{sp} 
&= \left\|  V^\T \hat{\Sigma}_{x} V -V^\T \Sigma_{x} V        \right\|_\mathrm{sp} \\
&= \left\|  V^\T_{\mathcal{S}_v} (\hat{\Sigma}_{x,\mathcal{S}_v \times \mathcal{S}_v} - \Sigma_{x,\mathcal{S}_v \times \mathcal{S}_v}) V_{\mathcal{S}_v}        \right\|_\mathrm{sp}  \\
&\le \left\|  \Sigma^{1/2}_{x,\mathcal{S}_v\times \mathcal{S}_v} V_{\mathcal{S}_v} \right\|_\mathrm{sp} ^2 
 \left\|  \Sigma^{-1/2}_{x,\mathcal{S}_v\times \mathcal{S}_v}  \hat{\Sigma}_{x,\mathcal{S}_v\times \mathcal{S}_v}  \Sigma^{-1/2}_{x,\mathcal{S}_v\times \mathcal{S}_v} - I_s                      \right\|_\mathrm{sp} \\
&\le \left\|  \Sigma^{-1/2}_{x,\mathcal{S}_v\times \mathcal{S}_v}  \hat{\Sigma}_{x,\mathcal{S}_v\times \mathcal{S}_v}  \Sigma^{-1/2}_{x,\mathcal{S}_v\times \mathcal{S}_v} - I_s\right\|_\mathrm{sp} ,
\end{split}
\end{equation}
where the last inequality follows from the fact that $\left\|\Sigma^{1/2}_{x,\mathcal{S}_v\times \mathcal{S}_v} V_{\mathcal{S}_v} \right\|_\mathrm{sp}   \le 1$.

It remains to show that  $ \left\|  \Sigma^{-1/2}_{x,\mathcal{S}_v\times \mathcal{S}_v}  \hat{\Sigma}_{x,\mathcal{S}_v\times \mathcal{S}_v}  \Sigma^{-1/2}_{x,\mathcal{S}_v\times \mathcal{S}_v} - I_s\right\|_\mathrm{sp} $ is upper bounded with high probability.
Let ${Z} = {X}_{\mathcal{S}_v}    \Sigma^{-1/2}_{x,\mathcal{S}_v\times \mathcal{S}_v}    \in \mathbb{R}^{n \times s}$.  Thus, each row of ${Z}$ is independent isotropic sub-Gaussian random variables.
Therefore, by an application of Proposition~\ref{prop:zmm}, 
\begin{equation*}
\begin{split}
&\mathrm{pr}\left\{  \left\|  \Sigma^{-1/2}_{x,\mathcal{S}_v\times \mathcal{S}_v}  \hat{\Sigma}_{x,\mathcal{S}_v\times \mathcal{S}_v}  \Sigma^{-1/2}_{x,\mathcal{S}_v\times \mathcal{S}_v} - I_s\right\|_\mathrm{sp}  \le\max(\delta,\delta^2) \right\}\\
&=   \mathrm{pr}\left\{ \left\|\frac{1}{n} Z^\T Z  - I_s\right\|_\mathrm{sp}  \le\max(\delta,\delta^2) \right\} \\
&\ge 1- 2   \exp(-ct^2/2).
 \end{split}
 \end{equation*}

 Picking $t=C s^{1/2}$ for sufficiently large $C>0$, we have 
 \[
  \left\|  \Sigma^{-1/2}_{x,\mathcal{S}_v\times \mathcal{S}_v}  \hat{\Sigma}_{x,\mathcal{S}_v\times \mathcal{S}_v}  \Sigma^{-1/2}_{x,\mathcal{S}_v\times \mathcal{S}_v} - I_s\right\|_\mathrm{sp}  \le C (s/n)^{1/2}
  \]
 with probability at least $1-\exp (   -C' s )$.  Substituting the last expression into (\ref{Eq:lemmadifference2}) and combining (\ref{Eq:lemmadifference1-5}) and (\ref{Eq:lemmadifference2}), we obtain 
 \begin{equation}
 \label{Eq:lemmadifference3}
  \left\| ( V^\T \hat{\Sigma}_{x} V )^{1/2}-I_K    \right\|_\mathrm{sp}  \le C (s/n)^{1/2},
 \end{equation}
with probability at least $1-\exp (  -C's )$.

It remains to obtain an upper bound for $\| (V^\T \hat{\Sigma}_{x}V)^{1/2}        \|_\mathrm{sp} $.
By Holder's inequality, we have 
\begin{equation*}
\begin{split}
\left\| (V^\T \hat{\Sigma}_{x} V)^{1/2}        \right\|_\mathrm{sp}  &\le \|V\|_\mathrm{sp}  \|\hat{\Sigma}_{x,\mathcal{S}_v\times \mathcal{S}_v}\|_\mathrm{sp} ^{1/2}\\
& \le c^{1/2} \left( \|\hat{\Sigma}_{x,\mathcal{S}_v\times \mathcal{S}_v}-{\Sigma}_{x,\mathcal{S}_v\times \mathcal{S}_v}\|_\mathrm{sp} ^{1/2}+\|{\Sigma}_{x,\mathcal{S}_v\times \mathcal{S}_v}\|_\mathrm{sp} ^{1/2}\right)\\
&\le  c^{1/2}C (s/n)^{1/2}+ c,
\end{split}
\end{equation*}
with probability at least $1-\exp (-C' s)$, where the last inequality is obtained by an application of  Lemma~\ref{lemma:sample eigenvalue} with fixed support $\mathcal{S}_v$.
Thus, from (\ref{Eq:lemmadifference1}), we obtain 
\[
\|\tilde{\Lambda}-\Lambda\|_\mathrm{sp}  \le \|\Lambda\|_\mathrm{sp}  C (s/n)^{1/2} \left( 1 + \| (V^\T \hat{\Sigma}_{x}V)^{1/2}        \|_\mathrm{sp} \right) \le C' (s/n)^{1/2}
\]
for sufficiently large $C'$, where we use the fact that $\|\Lambda\|_\mathrm{sp} $ is upper bounded by some constant.

To show the second part of the Lemma, using the definition of $\tilde{\Pi} $ and applying the Jensen's inequality, we have  
\begin{equation}
\small
\label{Eq:pi tilde}
\begin{split}
\|\tilde{\Pi}-\Pi\|_\mathrm{F} &= \left\|  V \left(V^\T \hat{\Sigma}_{x}  V\right)^{-1}V^\T -V V^\T  \right\|_\mathrm{F}\le \|V\|_\mathrm{sp}^2 \left\| I_K- \left(V^\T \hat{\Sigma}_{x}  V\right)^{-1}   \right\|_\mathrm{F}
\le cK \left\|I_K - \left( V^\T \hat{\Sigma}_{x}  V\right)^{-1} \right\|_\mathrm{sp},
\end{split}
\end{equation}
where the second inequality holds using the fact that the Frobenius norm of a matrix is upper bounded by the rank $K$ times the operator norm of a matrix.  Thus, it suffices to obtain an upper bound for $ \|I_K - ( V^\T \hat{\Sigma}_{x}  V)^{-1} \|_\mathrm{sp}$.
First, note that 
\[
 \left\|I_K - (V^\T \hat{\Sigma}_{x}  V)^{-1} \right\|_\mathrm{sp}\le 
 \left\|V^\T \hat{\Sigma}_{x}  V - I_K \right\|_\mathrm{sp}  \left\|   ( V^\T \hat{\Sigma}_{x}  V)^{-1} \right\|_\mathrm{sp}.
\]
It remains to bound $\left\|   ( V^\T \hat{\Sigma}_{x}  V)^{-1} \right\|_\mathrm{sp}$. By Weyl's inequality, we have $\sigma_{\min}(V^\T {\Sigma}_x V) \le \sigma_{\min} (V^\T \hat{\Sigma}_x V) + \|V^\T \hat{\Sigma}_x V-V^\T {\Sigma}_x V \|_{\mathrm{sp}}$. Thus, with probability at least $1-\exp(C' s)$,
\begin{equation*}
\begin{split}
\left\|   ( V^\T \hat{\Sigma}_{x}  V)^{-1} \right\|_\mathrm{sp} &= \frac{1}{\sigma_{\min} \left(  V^\T \hat{\Sigma}_{x}  V \right)}\\
&\le \frac{1}{1-\|V^\T \hat{\Sigma}_x V - V^\T \Sigma_x V \|_{\mathrm{sp}}}\\
&\le C.
\end{split}
\end{equation*}
Combining the above terms, we have with probability at least $1-\exp(C' s)$,
\begin{equation}
\label{Eq:pi tilde2}
\begin{split}
 \left\|I_K - (V^\T \hat{\Sigma}_{x}  V)^{-1} \right\|_\mathrm{sp}&\le 
 \left\|V^\T \hat{\Sigma}_{x}  V - I_K \right\|_\mathrm{sp}  \left\|   ( V^\T \hat{\Sigma}_{x}  V)^{-1} \right\|_\mathrm{sp}\\
 &\le C  \left\|V^\T \hat{\Sigma}_{x}  V - I_K \right\|_\mathrm{sp} \\
 &\le C (s/n)^{1/2}, \\
\end{split}
\end{equation}
where the last inequality follows from (\ref{Eq:lemmadifference3})  and Lemma~\ref{lemma:sample eigenvalue} with fixed support.
Substituting (\ref{Eq:pi tilde2}) into (\ref{Eq:pi tilde}), we obtain the desired results.

\end{proof}

%%%%%%%%%%%%%%%%%%%%%%%%
% Proof of Lemma curvature
%%%%%%%%%%%%%%%%%%%%%%%%
\subsection{Proof of Lemma~\ref{lemma:curvature}}
\begin{proof}
Let $u_j$ be the $j$th column of $U$ and let $a_j = u_j^\T E u_j$.
The term 
\[
\frac{d_K}{2}  \bigl\|UU^\T - E \bigr\|_\mathrm{F}^2
\]
can be upper bounded as 
\begin{equation}
\label{Eq:lemma:curvature1}
\begin{split}
\frac{d_K}{2}  \bigl\| UU^\T -E \bigr\|_\mathrm{F}^2 &= \frac{d_K}{2} \bigl\{
\bigl\|UU^\T \bigr\|_\mathrm{F}^2 + \bigl\| E\bigr\|_\mathrm{F}^2 - 2 \mathrm{tr} \bigl(U^\T E  U   \bigr)
   \bigr\}\\
   &\le \frac{d_K}{2} \bigl\{  
   \mathrm{tr} \bigl( I_K    \bigr) + \|E\|_\mathrm{sp} \|E\|_* -2 \mathrm{tr} \bigl(U^\T E U \bigr)
   \bigr\}\\
      &\le d_K \left(  
K - \sum_{j=1}^K a_j
   \right)\\
   &= d_K \sum_{j=1}^K \left( 1-a_j \right).
\end{split}
\end{equation}
Moreover, we have 
\begin{equation}
\label{Eq:lemma:curvature2}
\begin{split}
\bigl\langle ULU^\T, UU^\T -E \bigr\rangle  &=
\bigl\langle UDU^\T, UU^\T - E \bigr\rangle +\bigl\langle U (L-D) U^\T, UU^\T - E \bigr\rangle \\
&\ge\mathrm{tr} \bigl( UDU^\T- UDU^\T E   \bigr) 
- \| L-D  \|_\mathrm{F} \| UU^\T -E\|_\mathrm{F}\\
&=\mathrm{tr} \bigl(  D (I_K  - U^\T E U )  \bigr) - \| L-D  \|_\mathrm{F} \| UU^\T -E\|_\mathrm{F}\\
&\ge d_K \sum_{j=1}^K (1-a_j) - \| L-D  \|_\mathrm{F} \| UU^\T -E\|_\mathrm{F}.
\end{split}
\end{equation}
Combining (\ref{Eq:lemma:curvature1}) and (\ref{Eq:lemma:curvature2}), we have the desired result.

\end{proof}

%%%%%%%%%%%%%%%%%%%%%%%%
% Proof of Lemma optim
%%%%%%%%%%%%%%%%%%%%%%%%
\subsection{Proof of Lemma~\ref{lemma:optim}}
\begin{proof}
It suffices to show that $\tilde{\Pi}$  satisfies the constrains  $\bigl\| \hat{\Sigma}_{x}^{1/2} \tilde{\Pi} \hat{\Sigma}^{1/2}_{x} \bigr\|_* \leq K$ and $\bigl\| \hat{\Sigma}^{1/2}_{x} \tilde{\Pi} \hat{\Sigma}^{1/2}_{x} \bigr\|_\mathrm{sp}\leq 1$.  
By the definition of $\tilde{V}$, we have
\[
\tilde{V}^\T  \hat{\Sigma}_{x} \tilde{V}  = 
( V^\T \hat{\Sigma}_{x} V   )^{-1/2} V^\T \hat{\Sigma}_{x} V
  (V^\T \hat{\Sigma}_{x} V   )^{-1/2} = I_{K}.
\]
This implies that $\hat{\Sigma}^{1/2}_{x} \tilde{V}$ is an orthogonal matrix.  Thus, 
\begin{equation}
\label{Eq:lemma:optim1}
\|\hat{\Sigma}^{1/2}_{x} \tilde{\Pi} \hat{\Sigma}^{1/2}_{x}\|_\mathrm{sp} \le  \|\hat{\Sigma}^{1/2}_{x} \tilde{V}\|_\mathrm{sp}^2 = 1.
\end{equation}
Moreover, we have 
\begin{equation}
\label{Eq:lemma:optim2}
\mathrm{tr} \left( \hat{\Sigma}^{1/2}_{x} \tilde{\Pi} \hat{\Sigma}^{1/2}_{x} \hat{\Sigma}^{1/2}_{x} \tilde{\Pi} \hat{\Sigma}^{1/2}_{x}  \right)
= \mathrm{tr} \left\{ \hat{\Sigma}^{1/2}_{x} V (V^\T \hat{\Sigma}_{x} V)^{-1}  V^\T \hat{\Sigma}^{1/2}_{x}  \right\} = K.
\end{equation}
Combining (\ref{Eq:lemma:optim1}) and (\ref{Eq:lemma:optim2}), we have 
$\bigl\| \hat{\Sigma}^{1/2}_{x} \tilde{\Pi} \hat{\Sigma}^{1/2}_{x} \bigr\|_* =K$ and $\bigl\| \hat{\Sigma}^{1/2}_{x} \tilde{\Pi} \hat{\Sigma}^{1/2}_{x} \bigr\|_\mathrm{sp} \le 1$.
\end{proof}

%%%%%%%%%%%%%%%%%%%%%%%%
% Proof of Lemma complement set
%%%%%%%%%%%%%%%%%%%%%%%%
\subsection{Proof of Lemma~\ref{lemma:complement set}}
\begin{proof}
By the definition of the sets $\mathcal{S}_1^c,\ldots,\mathcal{S}_J^c$, we have 
\begin{equation}
 \label{Eq:lemma:complement set1}
l\cdot  \|\Delta_{\mathcal{S}_{j}^c}\|_{\max}\le\|\Delta_{\mathcal{S}_{j-1}^c}\|_1,
\end{equation}
since all $l$ elements of $\Delta_{\mathcal{S}_{j-1}^c}$ are larger than the elements of  $\Delta_{\mathcal{S}_{j}^c}$. Thus, we have
\begin{equation*}
\sum_{j=2}^J \|\Delta_{\mathcal{S}_j^c}\|_\mathrm{F} \le l^{1/2} \sum_{j=2}^J \|\Delta_{\mathcal{S}_j^c}\|_{\max} \le l^{-1/2} \sum_{j=2}^J\|  \Delta_{\mathcal{S}_{j-1}^c}\|_1\le l^{-1/2}  \| \Delta_{\mathcal{S}^c}  \|_1,
\end{equation*}
where the second inequality holds by (\ref{Eq:lemma:complement set1}).
\end{proof}

%%%%%%%%%%%%%%%%%%%%%%%%
% Proof of Lemma optim
%%%%%%%%%%%%%%%%%%%%%%%%
\subsection{Proof of Lemma~\ref{lemma:sample eigenvalue}}
\begin{proof}
Let $\mathcal{T}\subset \{1,\ldots,d\}$ be a set with cardinality $|\mathcal{T}| = s$. We have  
\begin{equation}
 \label{Eq:lemma:sample eigenvalue0}
 \begin{split}
&\underset{\mathcal{T} \subset\{1,\ldots,d\}, |\mathcal{T}| = s}{\max}   \|\hat{\Sigma}_{x,\mathcal{T}\times \mathcal{T}}-\Sigma_{x,\mathcal{T}\times \mathcal{T}}\|_\mathrm{sp} \\
 &\le \underset{\mathcal{T} \subset\{1,\ldots,d\}, |\mathcal{T}| = s}{\max}  \|\Sigma_{x,\mathcal{T}\times \mathcal{T}}^{-1/2}\hat{\Sigma}_{x,\mathcal{T}\times \mathcal{T}}\Sigma_{x,\mathcal{T}\times \mathcal{T}}^{-1/2}-I_s\|_\mathrm{sp} \underset{\mathcal{T} \subset\{1,\ldots,d\}, |\mathcal{T}| = s}{\max}  \|\Sigma_{x,\mathcal{T}\times \mathcal{T}}\|_\mathrm{sp} \\
 &\le c \cdot  \underset{\mathcal{T} \subset\{1,\ldots,d\}, |\mathcal{T}| = s}{\max} \|\Sigma_{x,\mathcal{T}\times \mathcal{T}}^{-1/2}\hat{\Sigma}_{x,\mathcal{T}\times \mathcal{T}}\Sigma_{x,\mathcal{T}\times \mathcal{T}}^{-1/2}-I_s\|_\mathrm{sp},
 \end{split}
 \end{equation}
where the last inequality holds by Assumption~\ref{ass:bounded eigenvalues}. By a similar argument in the proof of Lemma~\ref{lemma:difference}, we have for any fixed set $\mathcal{T}$ with cardinality $|\mathcal{T}|=s$,
\begin{equation}
\label{lemma:sample eigenvalue1}
\mathrm{pr}\left\{  \left\|  \Sigma^{-1/2}_{x,\mathcal{T}\times \mathcal{T}}  \hat{\Sigma}_{x,\mathcal{T}\times \mathcal{T}}  \Sigma^{-1/2}_{x,\mathcal{T}\times \mathcal{T}} - I_s\right\|_\mathrm{sp} \ge \max(\delta,\delta^2) \right\}\le 2   \exp(-ct^2/2).
\end{equation}
Thus, by the union bound, we have 
\begin{equation*}
\begin{split}
&\mathrm{pr}\left\{\underset{\mathcal{T} \subset\{1,\ldots,d\}, |\mathcal{T}| = s}{\max}   \left\|  \Sigma^{-1/2}_{x,\mathcal{T}\times \mathcal{T}}  \hat{\Sigma}_{x,\mathcal{T}\times \mathcal{T}}  \Sigma^{-1/2}_{x,\mathcal{T}\times \mathcal{T}} - I_s\right\|_\mathrm{sp} \ge  \max(\delta,\delta^2)  \right\}\\
&\le \sum_{\mathcal{T} \subset\{1,\ldots,d\}, |\mathcal{T}| = s}\mathrm{pr}\left\{  \left\|  \Sigma^{-1/2}_{x,\mathcal{T}\times \mathcal{T}}  \hat{\Sigma}_{x,\mathcal{T}\times \mathcal{T}}  \Sigma^{-1/2}_{x,\mathcal{T}\times \mathcal{T}} - I_s\right\|_\mathrm{sp} \ge  \max(\delta,\delta^2) \right\}\\
&\le 2 {d\choose s} \exp(-ct^2/2)\\
&\le 2 \left(\frac{ed}{s}\right)^s \exp(-ct^2/2),
\end{split}
 \end{equation*}
where the first inequality follows from (\ref{lemma:sample eigenvalue1}).
Picking $t=C \{s\log (ed)\}^{1/2}$, we obtain 
\begin{equation}
\label{Eq:lemma:sample eigenvalue2}
\underset{\mathcal{T} \subset\{1,\ldots,d\}, |\mathcal{T}| = s}{\max}     \left\|  \Sigma^{-1/2}_{x,\mathcal{T}\times \mathcal{T}}  \hat{\Sigma}_{x,\mathcal{T}\times \mathcal{T}}  \Sigma^{-1/2}_{x,\mathcal{T}\times \mathcal{T}} - I_s\right\|_\mathrm{sp}\le  C \{s\log (ed)/n\}^{1/2}
\end{equation}
for sufficiently large constant $C$, with probability greater than $1- \exp \{C' s\log (ed)  \}$.  Thus, substituting (\ref{Eq:lemma:sample eigenvalue2}) into (\ref{Eq:lemma:sample eigenvalue0}), we have 
\begin{equation}
\label{Eq:lemma:sample eigenvalue3}
\lambda_{\max}(\hat{\Sigma}_{x},s)  \le \lambda_{\max}(\Sigma_{x},s) + cC \{s\log (ed)/n\}^{1/2} \le c +  cC  \{s\log (ed)/n\}^{1/2},
\end{equation}
where the last inequality holds by Assumption~\ref{ass:bounded eigenvalues}.

Using the same upper bound (\ref{Eq:lemma:sample eigenvalue2}), it can be shown that 
\[
\frac{1}{c} -  c  C \{s\log (ed)/n \}^{1/2} \le \lambda_{\min} (\hat{\Sigma}_{x},s),
\]
as desired.

\end{proof}

%%%%%%%%%%%%%%%%%%%%%%%%
% Proof of lemma:tilde conditional concentration
%%%%%%%%%%%%%%%%%%%%%%%%
\subsection{Proof of Lemma~\ref{lemma:tilde conditional concentration}}
\label{proof:lemma:tilde conditional concentration}
\begin{proof}
By the triangle inequality, we have 
\[
\|\hat{\Sigma}_{E(x \mid y)}- \tilde{\Sigma}_{E(x \mid y)}\|_{\max}
\le \|\hat{\Sigma}_{E(x \mid y)}- {\Sigma}_{E(x \mid y)}\|_{\max}+ 
\|\tilde{\Sigma}_{E(x \mid y)}- {\Sigma}_{E(x \mid y)}\|_{\max}.
\]
The first term can be upper bounded by an application of Corollary~\ref{cor:concentration}, i.e., $\|\hat{\Sigma}_{E(x \mid y)} - \Sigma_{E(x \mid y)}\|_{\max}\le  C(\log d/n)^{1/2}$ with probability at least $1-\exp (C' \log d)$.
It remains to show that the second term can be upper bounded by the same rate with high probability.  

By adding and subtracting terms, and the triangle inequality, we have 
\begin{equation}
\label{Eq:proof:lemma:tilde conditional concentration1}
\begin{split}
&\|\tilde{\Sigma}_{E(x \mid y)}- {\Sigma}_{E(x \mid y)}\|_{\max}\\
&\le 
\|(\hat{\Sigma}_{x}-\Sigma_{x}) V \Lambda V^\T \Sigma_{x}\|_{\max}+
\|\Sigma_{x} V \Lambda V^\T(\hat{\Sigma}_{x}-\Sigma_{x})\|_{\max}+
\|(\hat{\Sigma}_{x}-\Sigma_{x}) V \Lambda V^\T (\hat{\Sigma}_{x}-\Sigma_{x})\|_{\max}\\
&= I_1+I_2+I_3.
\end{split}
\end{equation}

We now obtain an upper bound for $I_1$.  Following the set of  arguments in the proof of Lemma 6.4 in \citet{gao2014sparse}, $I_1$ can be rewritten as 
\[
\underset{j,k}{\max}\left| \frac{1}{n}\sum_{i=1}^n [x_{ij}   (x_i^\T V \Lambda V^\T \Sigma_{x})_k- E\{x_{ij} (x_i^\T V \Lambda V^\T \Sigma_{x})_k \}]     \right|.
\]
Note that the each term inside the summand is independent and centered sub-exponential random variables with sub-exponential norm upper bounded by some constant.
By Lemma~\ref{lemma:bernstein} and applying union bound, we have that 
\[
I_1 \le C (\log d/n)^{1/2}
\]
with probability at least $1-\exp(-C'\log d)$. The term $I_2$ an be upper bounded in a similar fashion.

Finally, $I_3$ can be rewritten as 
\begin{equation*}
\begin{split}
I_3 &\le  \sum_{k=1}^K \lambda_k\|(\hat{\Sigma}_{x}-\Sigma_{x}) V_{k}  V_{k}^\T (\hat{\Sigma}_{x}-\Sigma_{x})\|_{\max}\\
&\le \sum_{k=1}^K \lambda_k   \max_{j,k}    
\left[  \frac{1}{n}\sum_{i=1}^n \{x_{ij}x_i^\T V_k - E(x_{ij}x_i^\T V_k) \}        \right]^2. 
\end{split}
\end{equation*}
However, note that the term 
\[
\left[  \frac{1}{n}\sum_{i=1}^n \{x_{ij}x_i^\T V_k - E(x_{ij}x_i^\T V_k) \}        \right]
\]
 is sub-exponential random variable with mean zero and sub-exponential norm upper bounded by a constant. Thus, by Lemma~\ref{lemma:bernstein}, we have 
\[
I_3 \le K  \lambda_1  C \frac{\log d}{n},
\]
with probability at least $1-\exp(C' \log d)$. Under the assumption that $K\lambda_1(\log d/n)^{1/2}$ is bounded by some constant, combining the upper bounds, we have 
\[
\|\hat{\Sigma}_{E(x \mid y)}-\tilde{\Sigma}_{E(x \mid y)} \|_{\max} \le C (\log d /n)^{1/2}
\]
as desired.

\end{proof}

%%%%%%%%%%%%%%%%%%%%
%%%%%%%%%%%%%%%%%%%%
% Some Auxiliary Lemmas
%%%%%%%%%%%%%%%%%%%%
%%%%%%%%%%%%%%%%%%%%
\section{Auxiliary Lemmas}
In this section, we provide some auxiliary lemmas that are used in our proofs. We first provide some definition of sub-Gaussian and sub-exponential random variables.  We refer the reader to Chapter 5 in an unpublished 2015 technical report, Vershynin, R. (arXiv:1011.3027) for details.   
Let $Z$ be a sub-Gaussian random variable.  Let $\|Z\|_{\psi_2}$ be the sub-Gaussian norm that takes the form 
\[
\|Z\|_{\psi_2}  = \underset{p\ge 1}{\sup} \;p^{-1/2} (E|Z|^p)^{1/p}.
\]
Similar, let $X$ be a sub-exponential random variable and let 
$\|X\|_{\psi_1}$  be the sub-exponential norm that takes the form
\[
\|X\|_{\psi_1}  = \underset{p\ge 1}{\sup} \;p^{-1/2} (E|X|^p)^{1/p}.
\]
We note that if $Z$ is centered normal random variable with variance $\sigma^2$, then $Z$ is sub-Gaussian with $\|Z\|_{\psi_2} \le C\sigma$.  
We provide the tail probability of sub-Gaussian random variable in the following lemma, which follows directly from Lemma 5.5 in an unpublished 2015 technical report, Vershynin, R. (arXiv:1011.3027)
\begin{lemma}
\label{lemma:tail subgaussian}
Let $X$ be a sub-Gaussian random variable with mean zero and 
sub-Gaussian norm $\|X\|_{\psi_2}\le L$ for some constant $L$.
Then, for all $t>0$, 
\[
\mathrm{pr}(|X|\ge t)\le \exp(1- C t^2 /L^2).
\]
\end{lemma}

The following lemma follows directly from Lemma 5.14 in an unpublished 2015 technical report, Vershynin, R. (arXiv:1011.3027), which summarizes the relationship between sub-Gaussian and sub-exponential random variables.  
\begin{lemma}
\label{lemma:subgaussian}
Let $Z_1$ and $Z_2$ be two sub-Gaussian random variables.  Then, $Z_1 Z_2$ is a sub-exponential random variable with 
\[
\|Z_1 Z_2\|_{\psi_1}\le C \max(\|Z_1\|_{\psi_2}^2,\|Z_2\|_{\psi_2}^2 ),
\]
where $C>0$ is a positive constant.
\end{lemma}

The following inequality is a Bernstein-type inequality for sub-exponential random variables in Proposition 5.16 in an unpublished 2015 technical report, Vershynin, R. (arXiv:1011.3027).

%%%%%%%%%%%%%%%%%%%%
% Bernstein 
%%%%%%%%%%%%%%%%%%%%
\begin{lemma}%[Proposition 5.16 in an unpublished 2015 technical report, Vershynin, R. (arXiv:1011.3027)]
\label{lemma:bernstein}
 Let $X_1,\ldots,X_n$ be independent centered sub-exponential random variables, and $K= \max_i \|X_i\|_{\psi_1}$.  Then, for any $t>0$, we have  
\[
\mathrm{pr}\left(\left| \frac{1}{n}  \sum_{i=1}^n X_i   \right| > t\right) \le 2 \exp \left\{  -C  \min \left( \frac{nt^2}{K^2}, \frac{nt}{K}  \right)\right\},
\]
where $C>0$ is an absolute constant.
\end{lemma}

\bibliography{reference}

\end{document}